\documentclass[12pt]{article}

\usepackage{scicite}
\usepackage{times}

\newenvironment{sciabstract}{%
\begin{quote} \bf}
{\end{quote}}

\usepackage{amsmath}
\usepackage{amsthm}  
\usepackage{amssymb}
\usepackage{color}
\usepackage{mathtools} 
\usepackage{multirow}
\usepackage{hyperref}
\usepackage{graphicx}
\usepackage{caption}
\usepackage[labelsep=colon]{caption}
\usepackage{mathpazo}
\usepackage{algorithm}
\usepackage{algorithmic}
\usepackage{booktabs}
\usepackage{xcolor}

\usepackage{array}

\makeatletter

\newcommand{\thickhline}{%
    \noalign {\ifnum 0=`}\fi \hrule height 1pt
    \futurelet \reserved@a \@xhline
}
\newcolumntype{"}{@{\hskip\tabcolsep\vrule width 1pt\hskip\tabcolsep}}
\makeatother

\renewcommand{\thetable}{\textbf{\arabic{table}}}
\renewcommand{\thefigure}{\textbf{\arabic{figure}}}

\begingroup
\catcode`\#=11

\endgroup

\newtheorem{lemma}{Lemma}

\newtheorem{theorem}{Theorem}

\newcommand{\RE}{\mathbb{R}}

\newcommand{\beginsupplement}{%
        \setcounter{table}{0}
        \renewcommand{\thetable}{S\arabic{table}}%
        \setcounter{algorithm}{0}
        \renewcommand{\thealgorithm}{S\arabic{algorithm}}%
        \setcounter{equation}{0}
        \renewcommand{\theequation}{S\arabic{equation}}%
        \setcounter{figure}{0}
        \renewcommand{\thefigure}{S\arabic{figure}}%
        \setcounter{section}{0}
        \renewcommand{\thesection}{S\arabic{section}}%
     }

\title{Closed-form Continuous-time Neural Networks}

\author
{Ramin Hasani~$^{1\star,*}$, Mathias Lechner~$^{2\star}$, Alexander Amini~$^1$,\\ Lucas Liebenwein~$^1$, Aaron Ray~$^1$, \\
Max Tschaikowski~$^3$, Gerald Teschl~$^4$, Daniela Rus~$^1$\\
\\
\normalsize{$^{1}$Massachusetts Institute of Technology (MIT), Cambridge, USA}\\
\normalsize{$^{2}$Institute of Science and Technology Austria (IST Austria), Austria}\\
\normalsize{$^{3}$Aalborg University, Denmark}\\
\normalsize{$^{4}$University of Vienna (Uni Wien), Austria}\\
\\
\normalsize{$^{\star}$These authors contributed equally to the paper}\\
\normalsize{$^*$To whom correspondence should be addressed; E-mail:  rhasani@mit.edu.}
}

\date{}

\begin{document} 


\baselineskip24pt


\maketitle

\begin{sciabstract}
Continuous-time neural processes are performant sequential decision-makers that are built by differential equations (DE). However, their expressive power when they are deployed on computers is bottlenecked by numerical DE solvers. This limitation has significantly slowed down scaling and understanding of numerous natural physical phenomena such as the dynamics of nervous systems. Ideally we would circumvent this bottleneck by solving the given dynamical system in closed-form. This is known to be intractable in general. Here, we show it is possible to closely approximate the interaction between neurons and synapses -- the building blocks of natural and artificial neural networks -- constructed by liquid time-constant networks (LTCs) \cite{hasani2021liquid} efficiently in closed-form. 
To this end, we compute a tightly-bounded approximation of the solution of an integral appearing in LTCs' dynamics, that has had no known closed-form solution so far. This closed-form solution substantially impacts the design of continuous-time and continuous-depth neural models; for instance, since time appears explicitly in closed-form, the formulation relaxes the need for complex numerical solvers. Consequently, we obtain models that are between one and five orders of magnitude faster in training and inference compared to differential equation-based counterparts. More importantly, in contrast to ODE-based continuous networks, closed-form networks can scale remarkably well compared to other deep learning instances. Lastly, as these models are derived from liquid networks, they show remarkable performance in time series modeling, compared to advanced recurrent models.
\end{sciabstract}

\noindent \textbf{One Sentence Summary:} We find an approximate closed-form solution for the interaction of neurons and synapses and build a strong artificial neural network model out of it.

\noindent \textbf{Main Text:}

\noindent Continuous neural network architectures built by ordinary differential equations (ODEs) \cite{chen2018neural} opened a new paradigm for obtaining expressive and performant neural models. These models transform the depth dimension of static neural networks and the time dimension of recurrent neural networks into a continuous vector field, enabling parameter sharing, adaptive computations, and function approximation for non-uniformly sampled data.

These continuous-depth (time) models have shown promise in density estimation applications \cite{grathwohl2018ffjord,dupont2019augmented,yang2019pointflow,liebenwein2021sparse}, as well as modeling sequential and irregularly-sampled data \cite{rubanova2019latent,gholami2019anode,lechner2020learning,hasani2021liquid}. 

While ODE-based neural networks with careful memory and gradient propagation design \cite{lechner2020learning} perform competitively with advanced discretized recurrent models on relatively small benchmarks, their training and inference are slow due to the use  of advanced numerical DE solvers \cite{prince1981high}. This becomes even more troublesome as the complexity of the data, task and state-space increases (i.e., requiring more precision) \cite{raissi2019physics}, for instance, in open-world problems such as medical data processing, self-driving cars, financial time-series, and physics simulations.

The research community has developed solutions for resolving this computational overhead and for facilitating the training of neural ODEs, for instance, by relaxing the stiffness of a flow by state augmentation techniques \cite{dupont2019augmented,massaroli2020dissecting}, reformulating the forward-pass as a root-finding problem \cite{bai2019deep}, using regularization schemes \cite{finlay2020train,massaroli2020stable,kidger2020hey}, or improving the inference time of the network \cite{poli2020hypersolvers}.

\begin{figure}[t]
	\centering
	\includegraphics[width=1\textwidth]{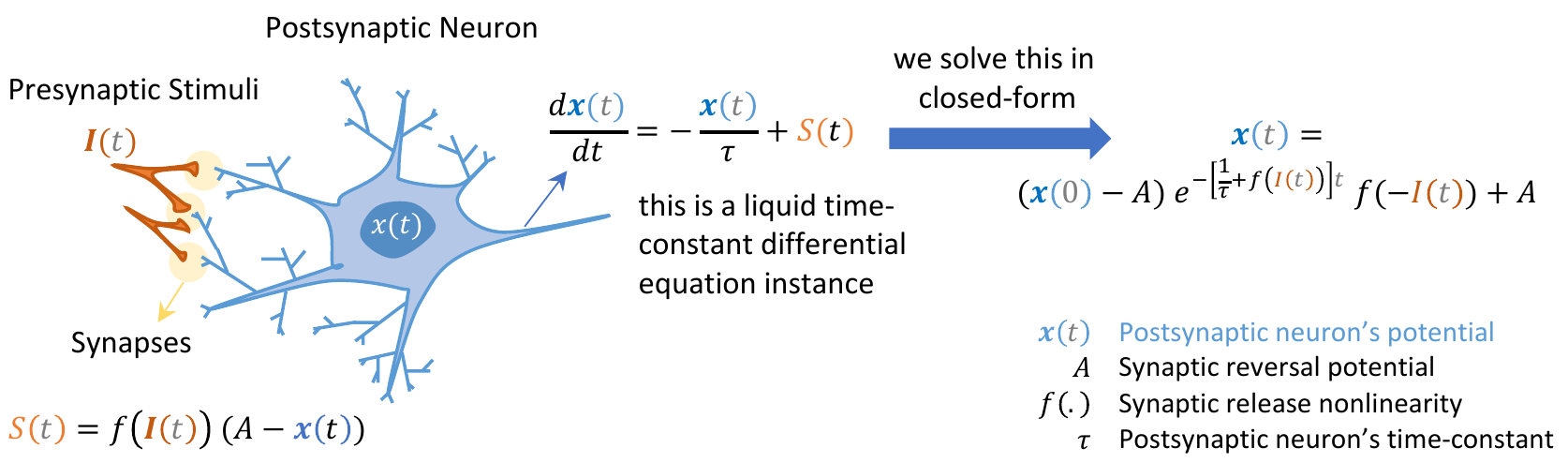}
	\caption{\textbf{Neural and Synapse Dynamics.} A postsynaptic neuron receives stimuli $I(t)$, through a nonlinear conductance-based synapse model. The dynamics of the membrane potential of this postsynaptic neuron is given by the differential equation presented in the middle. This equation is a fundamental building block of liquid time-constant networks (LTCs) \cite{hasani2021liquid}, for which there is no known closed-form expression. Here, we provided an approximate solution for this equation which shows the interaction of nonlinear synapses with a postsynaptic neurons, in closed-form.}
	\label{fig:neural_representation}
\end{figure}

In this paper, we take a step back and propose a fundamental solution: we derive a closed-form continuous-depth model that has the rich modeling capabilities of ODE-based models and does not require any solver to model data (see Figure \ref{fig:neural_representation}). The proposed continuous neural networks yield significantly faster training and inference speeds while being as expressive as their ODE-based counterparts. We provide a derivation for the approximate closed-form solution to a class of continuous neural networks that explicitly models time. We demonstrate how this transformation can be formulated into a novel neural model and scaled to create flexible, highly performant and fast neural architectures on challenging sequential datasets.

\noindent \textbf{Deriving an Approximate Closed-form Solution for Neural Interactions.} Two neurons interact with each other through synapses as shown in Figure \ref{fig:neural_representation}. There are three principal mechanisms for information propagation in natural brains that are abstracted away in the current building blocks of deep learning systems: 1) neural dynamics are typically continuous processes described by differential equations (c.f., dynamics of $x(t)$ in Figure \ref{fig:neural_representation}), 2) synaptic release is much more than scalar weights; it involves a nonlinear transmission of neurotransmitters, the probability of activation of receptors, and the concentration of available neurotransmitters, among other nonlinearities (c.f., $S(t)$ in Figure \ref{fig:neural_representation}), and 3) the propagation of information between neurons is induced by feedback and memory apparatuses (c.f. $I(t)$ stimulates $x(t)$ through a nonlinear synapse $S(t)$ which also has a multiplicative difference of potential to the postsynaptic neuron accounting for a negative feedback mechanism). Liquid time-constant (LTC) networks \cite{hasani2021liquid}, which are expressive continuous-depth models obtained by a bilinear approximation \cite{friston2003dynamic} of neural ODE formulation \cite{chen2018neural} are designed based on these mechanisms. Correspondingly, we take their ODE semantics and approximate a closed-form solution for the scalar case of a postsynaptic neuron receiving an input stimuli from a presynaptic source through a nonlinear synapse. 

To this end, we apply the theory of linear ODEs \cite{PerkoODEs} to analytically solve the dynamics of an LTC differential equation shown in Figure \ref{fig:neural_representation}. We then simplify the solution to the point where there is one integral left to solve. This integral compartment, $\int_0^t f(I(s)) ds$ in which $f$ is a positive, continuous, monotonically increasing, and bounded nonlinearity, is challenging to solve in closed-form; since it has dependencies on an input signal $I(s)$ that is arbitrarily defined (such as a real-world sensory readouts). To approach this problem, we discretize $I(s)$ into piecewise constant segments and obtain the discrete approximation of the integral in terms of sum of piecewise constant compartments over intervals. This piecewise constant approximation inspired us to introduce an approximate closed-form solution for the integral $\int_0^t f(I(s)) ds$ that is provably tight when the integral appears as the exponent of an exponential decay, which is the case for LTCs. We theoretically justify how this closed-form solution represents LTCs' ODE semantics and is as expressive (see Figure \ref{fig:neural_representation}). 

\noindent\textbf{Explicit Time Dependency.} We then dissect the properties of the obtained closed-form solution and design a new class of neural network models we call Closed-form Continuous-depth networks (CfC). CfCs have an explicit time dependency in their formulation that does not require an ODE solver to obtain their temporal rollouts. Thus, they maximize the trade-off between accuracy and efficiency of solvers (See Table \ref{tab:1}). CfCs perform computations at least one order of magnitude \textit{faster training and inference time} compared to their ODE-based counterparts, without loss of accuracy.

\begin{table}[t]
	\caption{\textbf{Time Complexity of the process to compute $K$ solver's steps.} $\epsilon$ is step-size, $\tilde \epsilon$ is the max step-size and $\delta << 0$. $\tilde K$ is time steps for closed-form continuous depth models (CfCs) which is equivalent to K. Table is reproduced and taken from \cite{poli2020hypersolvers}.}
	\setlength{\tabcolsep}{4pt}
	\centering
	\begin{tabular}{lll}
		\toprule
		\multicolumn{1}{c}{Method} & \multicolumn{1}{c}{Complexity} & Local Error \\
		\midrule
		$p$-th order solver & $\mathcal{O}(K\cdot p)$ & $\mathcal{O}(\epsilon^{p+1})$ \\
		adaptive--step solver & $-$  & $\mathcal{O}(\tilde \epsilon^{~p+1})$\\
		Euler {\tt hypersolver}& $ \mathcal{O}(K)$ & $\mathcal{O}(\delta \epsilon^{2})$ \\
		$p$-th order {\tt hypersolver} & $ \mathcal{O}(K\cdot p)$ & $\mathcal{O}(\delta \epsilon^{p+1})$ \\
		CfC (Ours) & $\mathcal{O}(\tilde K)$ & \textbf{not relevant} \\
		\bottomrule
	\end{tabular}
	\label{tab:1}
\end{table}

\begin{table}[t]
\centering
\caption{\textbf{Sequence and time-step prediction complexity.} $n$ is the sequence length, $k$ the number of hidden units, and $p$ = order of the ODE-solver.}
    \begin{tabular}{lll}\toprule
         Model & Sequence & Time-step  \\
          & prediction & prediction \\
         \midrule
         RNN & $\mathcal{O}(nk)$ &  $\mathcal{O}(k)$\\
         ODE-RNN & $\mathcal{O}(nkp)$ & $\mathcal{O}(kp)$\\
         Transformer & $\mathcal{O}(n^2 k)$ & $\mathcal{O}(nk)$\\
         CfC & $\mathcal{O}(nk)$ & $\mathcal{O}(k)$\\
         \bottomrule
    \end{tabular}
    \label{tab:transformers}
\end{table}

\noindent \textbf{Sequence and Time-step Prediction Efficiency.} CfCs perform per-time-step and per-sequence predictions by establishing a continuous flow similar to ODE-based models. However, as they do not require ODE-solvers, their complexity is at least one order of magnitude less than ODE based models. Consider  having a performant gated recurrent model \cite{hochreiter1997long} with the abilities to create expressive continuous flows \cite{chen2018neural} and adaptable dynamics \cite{hasani2021liquid}. Table \ref{tab:transformers} compares the time complexity of CfCs to that of standard RNNs, ODE-RNNs and Transformers. 

\noindent \textbf{CfCs: Flexible Deep Models for Sequential Tasks.} CfCs are equipped with novel gating mechanisms that explicitly control their memory. CfCs are as expressive as their ODE-based peers and can be supplied with mixed memory architectures \cite{lechner2020learning} to avoid gradient issues in sequential data processing applications. Beyond accuracy and performance metrics, our results indicate that when considering accuracy-per-compute time, CfCs exhibit over $150\times$ improvement. We perform a diverse set of advanced time series modeling experiments and present the performance and speed gain achievable by using CfCs in tasks with long-term dependencies, irregular data, and modeling physical dynamics, among others. 

\section*{Deriving a Closed-form Solution}
\label{sec:deriving}

In this section, we derive an approximate closed-form solution for liquid time-constant (LTC) networks, an expressive subclass of time-continuous models. We discuss how the scalar closed-form expression derived from a small LTC system can inspire the design of CfC models.

The hidden state of an LTC network is determined by the solution of the initial-value problem (IVP) given below \cite{hasani2021liquid}:
\begin{equation}
\label{eq:ltc}
    \frac{d\textbf{x}}{dt} = - (w_\tau + f(\textbf{x},\textbf{I},\theta)) \textbf{x}(t) + A f(\textbf{x},\textbf{I},\theta),
\end{equation}
where $\textbf{x}(t)$ defines the hidden states, \textbf{I}(t) is the input to the system, $w_\tau$ is a time-constant parameter vector, $A$ is a bias vector, and $f$ is a neural network parametrized by $\theta$.

\begin{theorem}
\label{theo:ltc_closed_form}
Given an LTC system determined by the IVP (\ref{eq:ltc}), constructed by one cell, receiving a single dimensional time-series input $I$ with no self connections, the following expression is an approximation of its closed-form solution: 
\begin{equation}
\label{eq:approx_closed_form_solution_ltc}
    x(t) = (x_0 - A) e^{-[w_\tau + f(I(t),\theta)]t} f(-I(t),\theta) + A
\end{equation}
\end{theorem}

\begin{proof}
In the single-dimensional case, the IVP (\ref{eq:ltc}) becomes linear in $x$ as follows:

\begin{align}
\frac{d}{dt} x (t) = - \big[w_\tau + f(I(t))\big] \cdot x(t) + A f(I(t))
\end{align}

\noindent Therefore, we can use the theory of linear ODEs to obtain an integral closed-form solution~ \cite[Section 1.10]{PerkoODEs} consisting of two nested integrals. The inner integral can be eliminated by means of integration by substitution~ \cite{Rudin76}. With this, the remaining integral expression can be solved in the case of piecewise constant inputs and approximated in the case of general inputs. The three steps of the proof are outlined below.

\noindent \textbf{Integral closed-form solution of LTC.} We consider the ODE semantics of a single neuron that receives some arbitrary continuous input signal $I$ and has a positive, bounded, continuous, and monotonically increasing nonlinearity $f$:
\begin{align*}
\frac{d}{dt} x (t) = - \big[w_\tau + f(I(t))\big] \cdot x(t) + A \cdot \big[w_\tau + f(I(t))\big]
\end{align*}

\noindent \textit{Assumption.} We assumed a second constant value $w_\tau$ in the above representation of a single LTC neuron. This is done to introduce symmetry on the structure of the ODE, hence being able to apply the theory of linear ODEs for solving the equation analytically.

\noindent By applying linear ODE systems theory~ \cite[Section 1.10]{PerkoODEs}, we obtain:
\begin{align}\label{eq_sol_double_int}
x(t) &= e^{- \int_0^t[w_\tau + f(I(s))] ds} \cdot x(0) + \nonumber \\ 
& \int_0^t e^{ - \int_s^t [w_\tau + f(I(v))] dv } \cdot A \cdot (w_\tau + f(I(s))) ds 
\end{align}

\noindent To resolve the double integral in the equation above, we define
\begin{align*}
u(s) & := \int_s^t [w_\tau + f(I(v))] dv,
\end{align*}
and observe that $\frac{d}{ds} u(s) = - (w_\tau + f(I(s)))$. Hence, integration by substitution allows us to rewrite~(\ref{eq_sol_double_int}) into:
\begin{align}
x(t) & = e^{- \int_0^t[w_\tau + f(I(s))] ds} \cdot x(0) - A \int_{u(0)}^{u(t)} e^{ - u } du \nonumber\\
& = x(0) e^{- \int_0^t[w_\tau + f(I(s))] ds} + A [ e^{-u} ]_{u(0)}^{u(t)} \nonumber\\
& = x(0) e^{- \int_0^t[w_\tau + f(I(s))] ds} + A \big( 1 - e^{- \int_0^t [w_\tau + f(I(s))] ds} \big) \nonumber\\
& = (x(0) - A) e^{-w_\tau t} e^{- \int_0^t f(I(s)) ds} + A \label{eq:ground:truth}
\end{align}

\noindent \textbf{Analytical LTC solution for piecewise constant inputs.} The derivation of a \emph{useful} closed-form expression of $x$ requires us to solve the integral expression $\int_0^t f(I(s)) ds$ for any $t \geq 0$. Fortunately, the integral $\int_0^t f(I(s)) ds$ enjoys a simple closed-form expression for piecewise constant inputs $I$. Specifically, assume that we are given a sequence of time points:
\[
0 = \tau_0 < \tau_1 < \tau_2 < \ldots < \tau_{n - 1} < \tau_n = \infty,
\]
such that $\tau_1,\ldots,\tau_{n-1} \in \RE$ and $I(t) = \gamma_i$ for all $t \in [\tau_i ; \tau_{i + 1})$ with $0 \leq i \leq n - 1$. Then, it holds that
\begin{equation}
\label{eq:discretized_int}
\int_0^t f(I(s)) ds =
f(\gamma_k) (t - \tau_k) + \sum_{i = 0}^{k-1} f(\gamma_i) (\tau_{i + 1} - \tau_i),
\end{equation}
when $\tau_k \leq t < \tau_{k + 1}$ for some $0 \leq k \leq n - 1$ (as usual, one defines $\sum_{i = 0}^{-1} := 0$). With this, we have:
\begin{align}
\label{eq:ltc_solution_in_discrete}
x(t) & = (x(0) - A) e^{-w_\tau t} e^{- f(\gamma_k) (t - \tau_k) - \sum_{i = 0}^{k-1} f(\gamma_i) (\tau_{i + 1} - \tau_i) } + A,
\end{align}
when $\tau_k \leq t < \tau_{k + 1}$ for some $0 \leq k \leq n - 1$. While any continuous input can be approximated arbitrarily well by a piecewise constant input~ \cite{Rudin76}, a tight approximation may require a large number of discretization points $\tau_1,\ldots,\tau_n$. We address this next.

\noindent \textbf{Analytical LTC approximation for general inputs.} Inspired by Eq.~\ref{eq:discretized_int}, the next result provides an analytical approximation of $x(t)$.

\begin{lemma}\label{lemma:approx}
For any Lipschitz continuous, positive, monotonically increasing, and bounded $f$ and continuous input signal $I(t)$, we approximate $x(t)$ in~(\ref{eq:ground:truth}) as follows:
\begin{align}
\tilde{x}(t) & = (x(0) - A) e^{-\big[ w_\tau t + f(I(t)) t \big]} f(-I(t)) + A \label{eq:approximation}
\end{align}
Then, $|x(t) - \tilde{x}(t)| \leq |x(0) - A| e^{-w_\tau t}$ for all $t \geq 0$. Writing $c = x(0) - A$ for convenience, we can obtain the following sharpness results, additionally:
\begin{enumerate}
    \item For any $t \geq 0$, we have $\sup\{ \tfrac{1}{c}(x(t) - \tilde{x}(t)) \mid I : [0;t] \to \RE \} = e^{-w_\tau t}$.
    \item For any $t \geq 0$, we have $\inf\{ \tfrac{1}{c}(x(t) - \tilde{x}(t)) \mid I : [0;t] \to \RE \} = e^{-w_\tau t} ( e^{-t} - 1)$.
\end{enumerate}
Above, the supremum and infimum are meant to be taken across all continuous input signals. These statements settle the question about the worst-case errors of the approximation. The first statement implies in particular that our bound is sharp.
\end{lemma}

\noindent The full proof is given in Methods. Lemma \ref{lemma:approx} demonstrates that the integral solution we obtained shown in Equation \ref{eq:ground:truth} is tightly close to the approximate closed-form solution we proposed in Equation \ref{eq:approximation}. Note that as $w_\tau$ is positively defined, the derived bound between Equations \ref{eq:ground:truth} and \ref{eq:approximation} ensures an exponentially decaying error as time goes by. 
\noindent Therefore, we have the statement of the theorem. 
\end{proof}

\begin{figure}[t]
	\centering
	\includegraphics[width=1\textwidth]{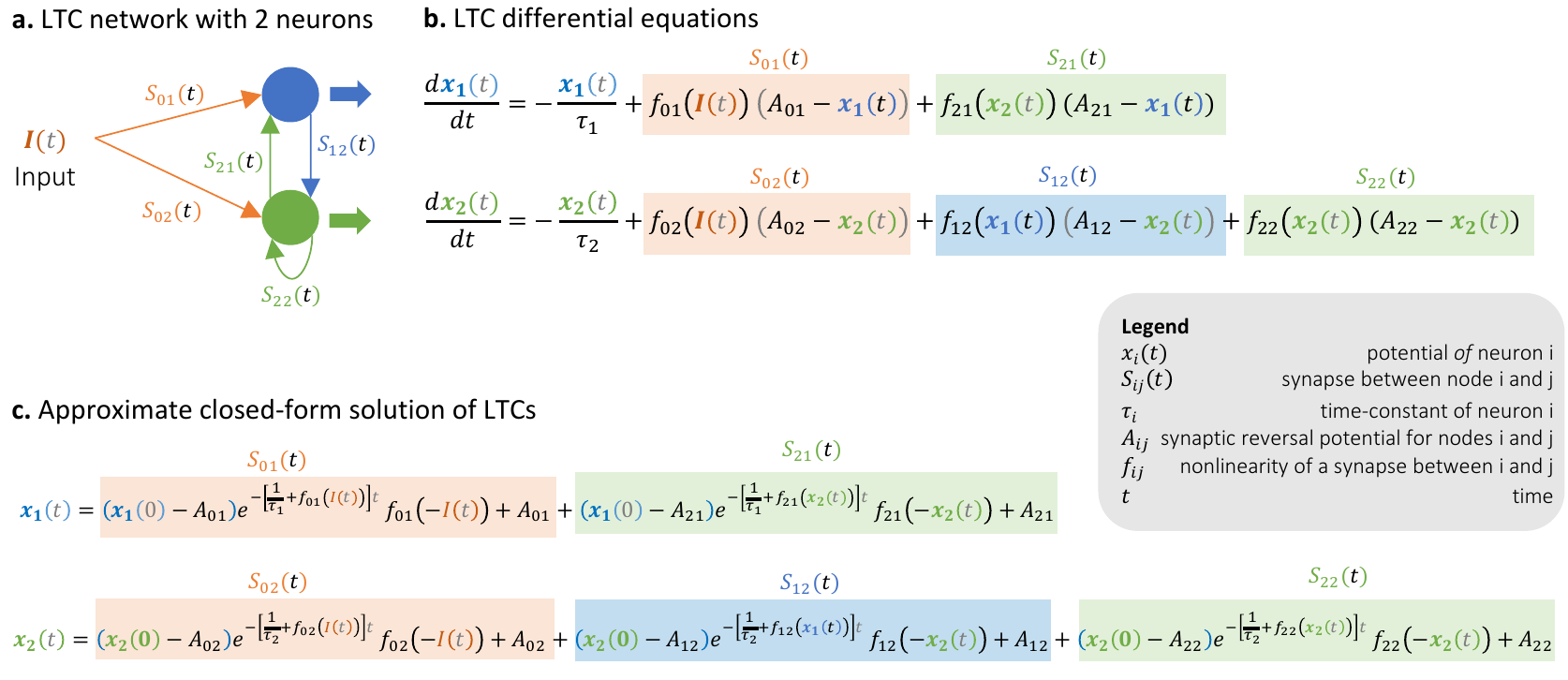}
	\caption{\textbf{Instantiation of LTCs in ODE and closed-form representations.} a) A sample LTC network with two nodes and five synapses. b) the ODE representation of this two-neuron system. c) the approximate closed-form representation of the network.}
	\label{fig:sample}
\end{figure}

\noindent \textbf{An Instantiation of LTCs and their approximate closed-form expressions.} 
Figure \ref{fig:sample} shows a liquid network with two neurons and five synaptic connections. The network receives an input signal I(t). Figure \ref{fig:sample} further derives the differential equation expression of the network along with its closed-form approximate solution. 

In general, it is possible to compile a trained LTC network into its closed-form version. This compilation allows us to speed up inference time of ODE-based networks as the closed-form variant does not require complex ODE solvers to compute outputs. Algorithm \ref{alg:compiler} provides the instructions on how to transfer a trained LTC network into its closed form variant.

\begin{algorithm}[t]
\caption{Translate a trained LTC network into its closed-form variant}
\label{alg:compiler}
\begin{algorithmic}
\STATE \textbf{Inputs:} LTC inputs $\textbf{I}^{(N \times T)}(t)$, LTC neurons activity $\textbf{x}^{(H \times T)}(t)$, and their initial states $\textbf{x}^{(H \times 1)}(0)$, Synapses adjacency matrix $W^{[(N+H) * (N+H)]}_{Adj}$ 
\STATE LTC's ODE Solver, Solver's step $\Delta t$, 
\STATE time-instance vectors of inputs, $\textbf{t}^{(1 \times T)}_{I(t)}$ 
\STATE time-instance of LTC neurons $\textbf{t}_{\textbf{x}(t)}$ \COMMENT{time might be sampled irregularly}
    \STATE LTC neurons' parameter $\tau^{(H \times 1)}$
    \STATE LTC network synaptic parameters \{ $\sigma^{(N \times H)}$, $\mu^{(N \times H)}$, $A^{(N \times H)}$\}
    \STATE \textbf{Outputs:} LTC's closed-form approximation of hidden state neurons, $\hat{\textbf{x}}^{(N \times T)}(t)$
\STATE $\textbf{x}_{pre}(t) = W_{Adj} \times [I_0 \dots I_N,~~~~~ x_0 \dots x_H]$ 
\COMMENT{ all presynaptic signals to nodes}
\FOR{ $i^{th}$ neuron in neurons 1 to $H$}
    \FOR{$j$ in Synapses to $i^{th}$ neuron}
        \STATE  \begin{equation}
        \hat{x}_{i} \mathrel{+}= (x_0 - A_{ij}) e^{\Big[-t_{x(t)} \odot \Big(1/\tau_i + \frac{1}{1+ e^{(-\sigma_{ij} (x_{pre_{ij}}-\mu_{ij}))}}\Big))\Big]} \odot \frac{1}{1+ e^{(\sigma_{ij} (x_{pre_{ij}}-\mu_{ij}))}} + A_{ij} \nonumber
            \end{equation}
    \ENDFOR
\ENDFOR
    \RETURN ${\hat{\textbf{x}}}(t)$
\end{algorithmic}
\end{algorithm}

\section*{Tightness of the Closed-form Solution in Practice}
\begin{figure}[t]
	\centering
	\includegraphics[width=0.7\textwidth]{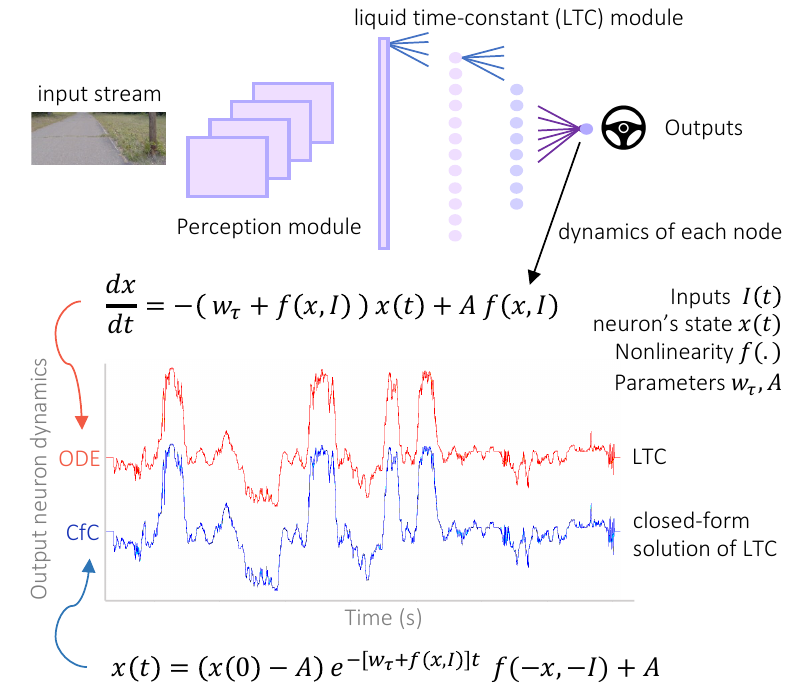}
	\caption{\textbf{Tightness of the closed-form solution in practice.} We approximate a closed-form solution for LTCs \cite{hasani2021liquid} while largely preserving the trajectories of their equivalent ODE systems. We develop our solution into closed-form continuous-depth (CfC) models that are at least 100x faster than neural ODEs at both training and inference on complex time-series prediction tasks.}
	\label{fig:intro}
\end{figure}

Figure \ref{fig:intro} shows an LTC-based network trained for autonomous driving \cite{lechner2020neural}. The figure further illustrates how close the proposed solution fits the actual dynamics exhibited from a single neuron ODE given the same parametrization. 

We took a trained Neural Circuit Policy (NCP)  \cite{lechner2020neural}, which consists of a perception module and a liquid time-constant (LTC) based network  \cite{hasani2021liquid} that possess 19 neurons and 253 synapses. The network was trained to autonomously steer a self-driving vehicle. We used recorded real-world test-runs of the vehicle for a lane-keeping task, governed by this network. The records included the inputs, outputs as well as all
LTC neurons’ activities and parameters. To perform a sanity check whether our proposed closed-form solution for LTC neurons is good enough in practice as well as the theory, we plugged in the parameters of individual neurons and synapses of the differential equations into the closed-form solution (Similar to the representations shown in Figure \ref{fig:sample}b and \ref{fig:sample}c) and emulated the structure of the ODE-based LTC networks. We then visualized the output neuron’s dynamics of the ODE (in blue) and of the closed-form
solution (in red). As illustrated in Figure \ref{fig:intro}, we observed that the behavior of the ODE is captured with a mean-squared error of 0.006 by the closed-form solution. This experiment is an empirical evidence for the tightness results presented in our theory. Hence, the closed-form solution contains the main properties of liquid networks in approximating dynamics. We next show how to design a novel neural network instance inspired by this closed-form solution, that has well-behaved gradient properties and approximation capabilities.


\section*{Design a Closed-form Continuous-depth Model Inspired by the Solution}

\begin{figure*}[t]
	\centering
	\includegraphics[width=1\textwidth]{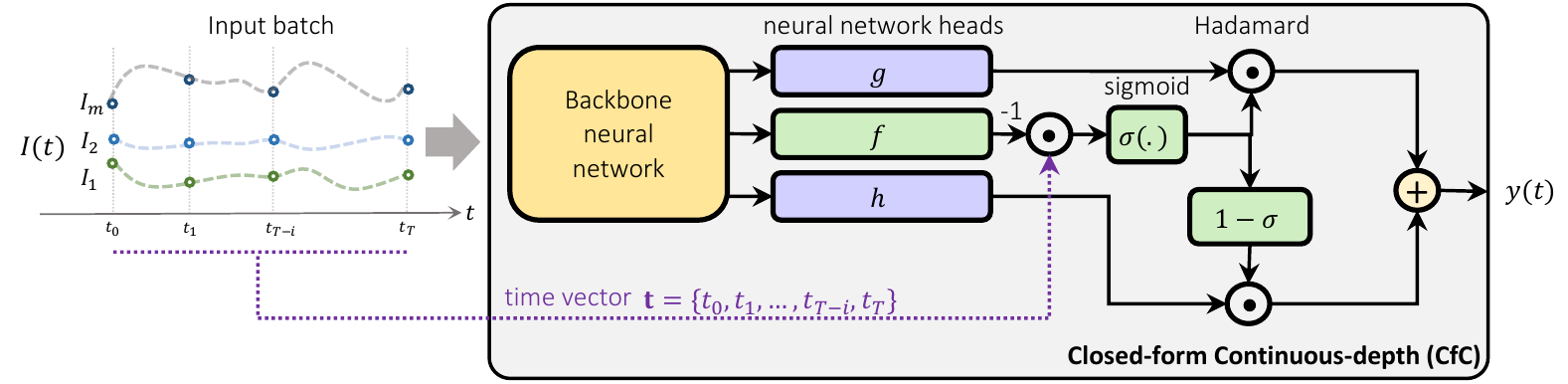}
	\caption{\textbf{Closed-form Continuous-depth neural architecture.} A baclbone neural network layer delivers the input signals into three head networks $g$, $f$ and $h$. $f$ acts as a \emph{liquid} time-constant for the sigmoidal time-gates of the network. $g$ and $h$ construct the nonlinearieties of the overall CfC network.}
	\label{fig:cfc}
\end{figure*}

Leveraging the scalar closed-form solution expressed by Eq. (\ref{eq:approx_closed_form_solution_ltc}), we can now distill this model into a neural network that can be trained at scale. The solution providing a grounded theoretical basis for solving scalar continuous-time dynamics and it is important to translate this theory into a practical neural network model which can be integrated into larger representation learning systems. Doing so requires careful attention to potential gradient and expressivity issues that can arise during optimization, which we will outline in this section. 

\noindent Formally, the hidden states, $\textbf{x}(t)^{(D \times 1)}$ with $D$ hidden units at each time step $t$, can be explicitly obtained by:

\begin{equation}
\label{eq:neural_network_cfc}
\textbf{x}(t) = B \odot e^{-[w_\tau + f(\textbf{x}, \textbf{I};\theta)] t} \odot f(-\textbf{x},-\textbf{I};\theta) + A,
\end{equation}
where $B^{(D)}$ collapses $(x_0-A)$ of Eq. \ref{eq:approx_closed_form_solution_ltc} into a parameter vector. $A^{(D)}$ and $w_{\tau}^{(D)}$ are system's parameter vectors, as well, $\textbf{I}(t)^{(m \times 1)}$ is an $m$-dimensional input at each time step $t$, $f$ is a neural network parametrized by $\theta = \{ W_{Ix}^{(m \times D)}, W_{xx}^{(D \times D)}, b_{x}^{(D)} \}$, and $\odot$ is the Hadamard (element-wise) product. 
While the neural network presented in \ref{eq:neural_network_cfc} can be proven to be a universal approximator as it is an approximation of an ODE system \cite{chen2018neural,hasani2021liquid}, in its current form, it has trainability issues which we point out and resolve shortly:

\noindent \textbf{Resolving the gradient issues.} The exponential term in Eq. \ref{eq:neural_network_cfc} derives the system's first part (exponentially fast) to 0 and the entire hidden state to $A$. This issue becomes more apparent when there are recurrent connections and causes vanishing gradient factors when trained by gradient descent \cite{hochreiter1991untersuchungen}. To reduce the effect, we replace the exponential decay term with a reversed sigmoidal nonlinearity $\sigma(.)$. This nonlinearity is approximately 1 at $t=0$ and approaches 0 in the limit $t\rightarrow \infty$. However, unlike the exponential decay, its transition happens much smoother, yielding a better condition on the loss surface.

\noindent \textbf{Replacing biases by learnable instances.} Next, we consider the bias parameter $B$ to be part of the trainable parameters of the neural network $f(-\textbf{x}, -\textbf{I}; \theta)$ and choose to use a new network instance instead of $f$ (presented in the exponential decay factor). We also replace $A$ with another neural network instance, $h(.)$ to enhance the flexibility of the model. To obtain a more general network architecture, we allow the nonlinearity $f(-\textbf{x}, -\textbf{I}; \theta)$ present in Eq. \ref{eq:neural_network_cfc} have both shared (backbone) and independent, ($g(.)$), network compartments.

\noindent \textbf{Gating balance.} The time-decaying sigmoidal term can play a gating role if we additionally multiply $h(.)$, with ($1 - \sigma(.)$). This way, the time-decaying sigmoid function stands for a gating mechanism that interpolates between the two limits of $t\rightarrow -\infty$ and $t\rightarrow \infty$ of the ODE trajectory. 

\noindent \textbf{Backbone.} Instead of learning all three neural network instances $f, g$ and $h$ separately, we have them share the first few layers in the form of a backbone that branches out into these three functions. As a result, the backbone allows our model to learn shared representations, thereby speeding up and stabilizing the learning process. More importantly, this architectural prior enables two simultaneous benefits: 1) Through the shared backbone a coupling between time-constant of the system and its state nonlinearity get established that exploits causal representation learning evident in a liquid neural network  \cite{hasani2021liquid,vorbach2021causal}. 2) through separate head network layers, the system has the ability to explore temporal and structural dependencies independently of each other.  

\noindent These modifications result in the closed-form continuous-depth (CfC) neural network model:

\begin{align}
\label{eq:cfc}
\textbf{x}(t) = \textcolor{black}{\underbrace{\sigma(-f(\textbf{x}, \textbf{I};\theta_f) ~\textcolor{black}{\textbf{t}})}_\text{time-continuous gating}} \odot g(\textbf{x},\textbf{I};\theta_g) + \\ \nonumber \textcolor{black}{\underbrace{\big[ 1 -\sigma(-[f(\textbf{x}, \textbf{I};\theta_f)] ~\textcolor{black}{\textbf{t}}) \big]}_\text{time-continuous gating}} \odot h(\textbf{x},\textbf{I};\theta_h).
\end{align}

The CfC architecture is illustrated in Figure \ref{fig:cfc}. The neural network instances could be selected arbitrarily. The time complexity of the algorithm is equivalent to that of discretized recurrent networks \cite{hasani2019response}, which is at least one order of magnitude faster than ODE-based networks. 

\noindent \textbf{How do you deal with time, $\textcolor{black}{\textbf{t}}$?} CfCs are continuous-depth models that can set their temporal behavior based on the task-under-test. For time-variant datasets (e.g., irregularly sampled time series, event-based data, and sparse data), $t$ for each incoming sample is set based on its time-stamp or order. For sequential applications where the time of the occurrence of a sample does not matter, $t$ is sampled \textit{batch-length}-times with equidistant intervals within two hyperparameters $a$ and $b$.

\section*{Experiments with CfCs}
We now assess the performance of CfCs in a series of sequential data processing tasks compared to advanced, recurrent models. We first evaluate how CfCs compare to LTC-based neural circuit policies (NCPs) \cite{lechner2020neural} in real-world autonomous lane keeping tasks. We then approach solving conventional sequential data modeling tasks (e.g., bit-stream prediction, sentiment analysis on text data, medical time-series prediction, and robot kinematics modeling), and compare CfC variants to an extensive set of advanced recurrent neural network baselines.

\noindent \textbf{CfC Network Variants.} To evaluate how the proposed modifications we applied to the closed-form solution network described by Eq. \ref{eq:neural_network_cfc}, we test four variants of the CfC architecture: 1) Closed-form solution network (\textbf{Cf-S}) obtained by Eq. \ref{eq:neural_network_cfc}, 2) CfC without the second gating mechanism (\textbf{CfC-noGate}). This variant does not have the $1-\sigma$ instance shown in Figure \ref{fig:cfc}. 3) Closed-form Continuous-depth model (\textbf{CfC}) expressed by Eq. \ref{eq:cfc}. 4) CfC wrapped inside a mixed-memory architecture (i.e., CfC defines the memory state of an RNN for instance an LSTM). We call this variant \textbf{CfC-mmRNN}. Each of these four proposed variants leverage our proposed solution, and thus are at least one order of magnitude faster than continuous-time ODE models.

\noindent \textbf{How well CfCs perform in autonomous driving compared to NCPs and other models?} In this experiment, our objective is to evaluate how robustly CfCs learn to perform autonomous navigation as opposed to its ODE-based counterparts LTC networks. The task is to map incoming pixel observations to steering curvature commands. We start off by training neural network architectures that possess a convolutional head stacked with the choice of RNN. The RNN compartment of networks are replaced by LSTM networks, NCPs \cite{lechner2020neural}, Cf-S, CfC-NoGate, and CfC-mmRNN. We also trained a fully convolutional neural network for the sake of proper comparison. 

Our training pipeline followed an imitation learning approach with paired pixel-control data, from a 30Hz BlackFly PGE-23S3C RGB camera, collected by a human expert driver across a variety of rural driving environments, including times of day, weather conditions, and season of the year. The original 3-hour dataset was further augmented to include off-orientation recovery data using a privileged controller\cite{amini2021vista} and a data-driven view synthesizer\cite{amini2020learning}. The privileged controller enabled training all networks using guided policy learning\cite{levine2013guided}. After training, all networks were transferred on-board our full-scale autonomous vehicle (Lexus RX450H, retrofitted with drive-by-wire capability). The vehicle was consistently started at the center of the lane, initialized with each trained model, and was run to completion at the end of the road. If the model exited the bounds of the lane a human safety driver intervened and restarted the model from the center of the road at the intervention location. All models were tested with and without noise added to the sensory inputs to evaluate robustness. 

\begin{figure}[t]
	\centering
	\includegraphics[width=0.8\textwidth]{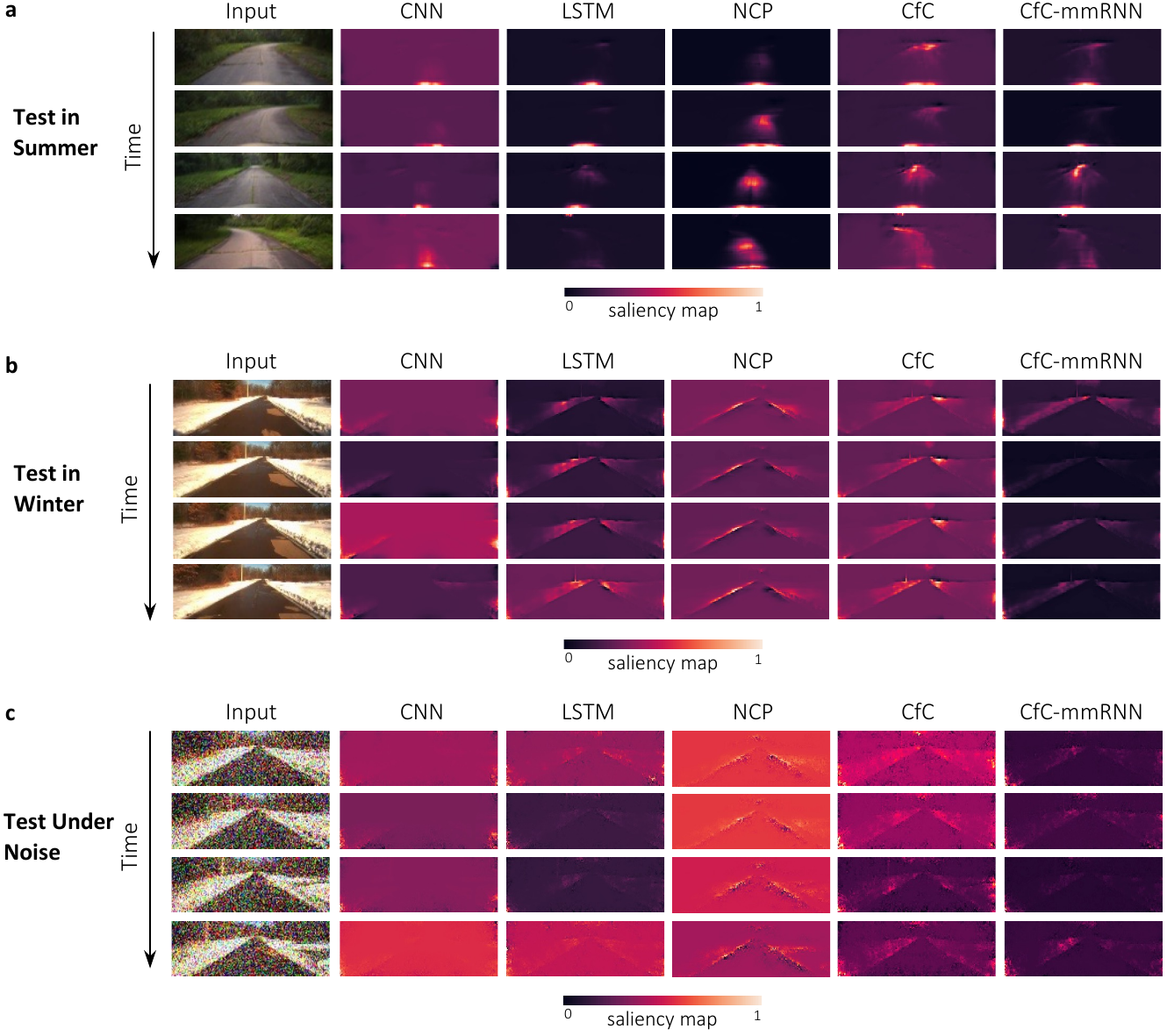}
	\caption{\textbf{Attention Profile of networks.} Trained networks receive unseen inputs (first column in each tab) and generate acceleration and steering commands. We use the Visual-Backprop algorithm \cite{bojarski2018visualbackprop} to compute the saliency maps of the convolutional part of each network. a) results for networks tested on data collected in summer. b) results for networks tested on data collected in winter. c) results for inputs corrupted by a zero-mean Gaussian noise with variance, $\sigma^2 = 0.35$.}
	\label{fig:saliency}
\end{figure}

The testing environment consisted of 1km of private test road with unlabeled lane-markers and we observed that all trained networks were able to successfully complete the lane-keeping task at a constant velocity of 30 km/hr. Fig. \ref{fig:saliency} provides an insight into how these networks come with driving decisions. To this end, we computed the attention of each network while driving, by using the visual-backprop algorithm\cite{bojarski2018visualbackprop}. We observe that CfCs similar to NCPs demonstrate a consistent attention pattern in each subtask, while maintaining their attention profile under heavy noise depicted in Fig. \ref{fig:saliency}c. Similar to NCPs, CfCs are very parameter efficient. They performed the end-to-end autonomous lane keeping task with around 4k trainable parameters in their recurrent neural network component.

In the following, we design sequence data processing pipelines where we extensively test CfCs' effectiveness in learning spatiotemporal dynamics, compared to a large range of advanced recurrent models. 

\begin{table}[t]
    \centering
        \caption{\textbf{Lane-keeping models' parameter count.} CfC and NCP networks perform lane-keeping in unseen scenarios with a compact representation.}
    \begin{tabular}{lcc}
    \toprule
        \textbf{Modes} &  \textbf{Total Parameter Count} & \textbf{RNN Parameter Count}\\
        & (CNN head + RNN) & \\
         \midrule
CNN & 2,124,901 & - \\
LSTM & 259,733 & 33089 \\
NCP & 233,139 & 6495 \\
Cf-S & 227,728 & 1084 \\
CfC & 230,828 & 4184\\
CfC-NoGate & 230,828 & 4184\\
CfC-mmRNN & 235,052 & 8408 \\
         \bottomrule
    \end{tabular}
    \label{tab:param_count}
\end{table}

\noindent \textbf{Baselines.}
We compare CfCs to a diverse set of advanced algorithms developed for sequence modeling by both discretized and continuous mechanisms. Examples include some variations of classical autoregressive RNNs, such as an RNN with concatenated $\Delta t$ (RNN-$\Delta t$), a recurrent model with moving average on missing values (RNN-impute), RNN Decay \cite{rubanova2019latent}, long short-term memory (LSTMs) \cite{hochreiter1997long}, and gated recurrent units (GRUs) \cite{chung2014empirical}. We also report results for a variety of encoder-decoder ODE-RNN based models, such as RNN-VAE, Latent variable models with RNNs, and with ODEs, all from \cite{rubanova2019latent}. 

Furthermore, we include models such as interpolation prediction networks (IP-Net) \cite{shukla2018interpolation}, Set functions for time-series (SeFT) \cite{horn2020set}, CT-RNNs \cite{funahashi1993approximation}, CT-GRU \cite{mozer2017discrete}, CT-LSTM \cite{mei2017neural}, GRU-D \cite{che2018recurrent}, Phased-LSTM \cite{neil2016phased}, bi-directional RNNs \cite{schuster1997bidirectional}. Finally, we benchmarked CfCs against competitive recent RNN architectures with the premise of tackling long-term dependencies, such as Legandre Memory Units (LMU) \cite{voelker2019legendre}, high-order polynomial projection operators (Hippo) \cite{gu2020hippo}, orthogonal recurrent models (expRNNs) \cite{lezcano2019cheap}, mixed memory RNNs such as (ODE-LSTMs) \cite{lechner2020learning}, coupled oscillatory RNNs (coRNN) \cite{rusch2021coupled}, and Lipschitz RNNs \cite{erichson2021lipschitz}. 

\subsection*{Regularly and Irregularly-Sampled Bit-Stream XOR}
The bit-stream XOR dataset \cite{lechner2020learning} considers classifying bit-streams implementing an XOR function in time, i.e., each item in the sequence contributes equally to the correct output. The bit-streams are provided in densely sampled and event-based sampled format. The densely sampled version simply represents an incoming bit as an input event. The event sampled version transmits only bit-changes to the network, i.e., multiple equal bit is packed into a single input event. Consequently, the densely sampled variant is a regular sequence classification problem, whereas the event-based encoding variant represents an irregularly sampled sequence classification problem. 

Table \ref{tab:synthetic} compares the performance of many RNN baselines. Many architectures such as Augmented LSTM, CT-GRU, GRU-D, ODE-LSTM, coRNN, and Lipschitz RNN, and all variants of CfC can successfully solve the task with 100\% accuracy when the bit-stream samples are equidistant from each other. However, when the bit-stream samples arrive at non-uniform distances, only architectures that are immune to the vanishing gradient in irregularly sampled data can solve the task. These include GRU-D, ODE-LSTM and CfCs, and CfC-mmRNNs. ODE-based RNNs cannot solve the event-based encoding tasks regardless of their choice of solvers, as they have vanishing/exploding gradient issues \cite{lechner2020learning}. The hyperparameter details of this experiment is provided in Table \ref{tab:hyperparamsxor}.

\begin{table}[t]
    \centering
    \footnotesize
    \caption{\textbf{Bit-stream XOR sequence classification.} The performance values for all baseline models are reproduced from \cite{lechner2020learning}. Numbers present mean $\pm$ standard deviations, n=5}
    \begin{tabular}{lllll}
    \toprule
        \multirow{2}{*}{Model} & Equidistant & Event-based & Time Per epoch  & ODE-based? \\
                               & encoding  & (irregular) encoding & (min) & \\
        \hline
        $\dagger$Augmented LSTM \cite{hochreiter1997long} & \textbf{100.00\%} $\pm$ 0.00 & 89.71\% $\pm$ 3.48 & 0.62 & No\\
        $\dagger$CT-GRU \cite{mozer2017discrete} & \textbf{100.00\%} $\pm$ 0.00 & 61.36\% $\pm$ 4.87 & 0.80 & No \\
        $\dagger$RNN Decay \cite{rubanova2019latent} & 60.28\% $\pm$ 19.87 & 75.53\% $\pm$ 5.28 & 0.90 & No \\
        $\dagger$Bi-directional RNN \cite{schuster1997bidirectional}  & \textbf{100.00\%} $\pm$ 0.00 & 90.17\% $\pm$ 0.69 & 1.82 & No \\
        $\dagger$GRU-D \cite{che2018recurrent} & \textbf{100.00\%} $\pm$ 0.00 & 97.90\% $\pm$ 1.71 & 0.58 & No \\
        $\dagger$PhasedLSTM \cite{neil2016phased}  & 50.99\% $\pm$ 0.76 & 80.29\% $\pm$ 0.99 & 1.22 & No \\
        $\dagger$CT-LSTM \cite{mei2017neural} & 97.73\% $\pm$ 0.08 & 95.09\% $\pm$ 0.30 & 0.86 & No \\
        coRNN \cite{rusch2021coupled} & \textbf{100.00\%} $\pm$ 0.00 & 52.89\% $\pm$ 1.25 & 0.57 & No \\
        Lipschitz RNN \cite{erichson2021lipschitz} & \textbf{100.00\%} $\pm$ 0.00 & 52.84\% $\pm$ 3.25 & 0.63 & No \\
        $\dagger$ODE-RNN \cite{rubanova2019latent}  & 50.47\% $\pm$ 0.06 & 51.21\% $\pm$ 0.37 & 4.11 & Yes\\
        $\dagger$CT-RNN \cite{funahashi1993approximation} & 50.42\% $\pm$ 0.12 & 50.79\% $\pm$ 0.34 & 4.83 & Yes\\
        $\dagger$GRU-ODE \cite{rubanova2019latent} & 50.41\% $\pm$ 0.40 & 52.52\% $\pm$ 0.35 & 1.55 & Yes \\
        $\dagger$ODE-LSTM \cite{lechner2020learning} & \textbf{100.00\%} $\pm$ 0.00 &
        \textbf{98.89\%} $\pm$ 0.26 & 1.18 & Yes \\
        LTC \cite{hasani2021liquid} & \textbf{100.00\%} $\pm$ 0.00 &  49.11\% $\pm$ 0.00 & 2.67 & Yes \\
        \cmidrule{1-5}
        Cf-S (ours) & \textbf{100.00\%} $\pm$ 0.00 &  85.42\% $\pm$ 2.84 & 0.36 & No \\
        CfC-noGate (ours) & \textbf{100.00\%} $\pm$ 0.00 & 96.29\% $\pm$ 1.61 & 0.78 & No \\
        CfC (ours) & \textbf{100.00\%} $\pm$ 0.00 & \textbf{99.42\%} $\pm$ 0.42 & 0.75 & No \\
        CfC-mmRNN (ours)  & \textbf{100.00\%} $\pm$ 0.00 & \textbf{99.72\%} $\pm$ 0.08 & 1.26 & No \\
        \bottomrule
    \end{tabular}
    \caption*{\footnotesize \textbf{Note:} The performance of models marked by $\dagger$ are reported from \cite{lechner2020learning}.}
    \label{tab:synthetic}
\end{table}


\subsection*{PhysioNet Challenge}
The PhysioNet Challenge 2012 dataset considers the prediction of the mortality of 8000 patients admitted to the intensive care unit (ICU). The features represent time series of medical measurements of the first 48 hours after admission. The data is irregularly sampled in time, and over features, i.e., only a subset of the 37 possible features is given at each time point.
We perform the same test-train split and preprocessing as \cite{rubanova2019latent}, and report the area under the curve (AUC) on the test set as metric in Table \ref{tab:physionet}. We observe that CfCs perform competitively to other baselines while performing 160 times faster training time compared to ODE-RNNs and 220 times compared to continuous latent models. CfCs are also, on average, three times faster than advanced discretized gated recurrent models. The hyperparameter details of this experiment is provided in Table \ref{tab:hyperparamsphysionet}. 

\begin{table}[t]
    \centering
    \caption{\textbf{PhysioNet}. The experiment is performed without any pretraining or pretrained word-embeddings. Thus, we excluded advanced attention-based models \cite{shukla2021multi,ad-trans2021} such as Transformers \cite{vaswani2017attention} and RNN structures that use pretraining. Numbers present mean $\pm$ standard deviations, n=5}
    \begin{tabular}{lcc}
    \toprule
         Model & AUC Score (\%) & time per epoch (min) \\\hline
         ${\dagger}$RNN-Impute \cite{rubanova2019latent}& 0.764 $\pm$ 0.016 & 0.5 \\
         ${\dagger}$RNN-delta-t \cite{rubanova2019latent} & 0.787 $\pm$ 0.014 & 0.5  \\
         ${\dagger}$RNN-Decay \cite{rubanova2019latent} & 0.807 $\pm$ 0.003 & 0.7 \\
         ${\dagger}$GRU-D \cite{che2018recurrent} & 0.818 $\pm$ 0.008 & 0.7 \\
         ${\dagger}$Phased-LSTM \cite{neil2016phased} & \textbf{0.836} $\pm$ 0.003 & 0.3 \\
         $\ast$IP-Nets \cite{shukla2018interpolation} & 0.819 $\pm$ 0.006 & 1.3 \\
         $\ast$SeFT \cite{horn2020set} & 0.795 $\pm$ 0.015 & 0.5 \\
         ${\dagger}$RNN-VAE \cite{rubanova2019latent} & 0.515 $\pm$ 0.040 & 2.0 \\
         ${\dagger}$ODE-RNN \cite{rubanova2019latent} & \textbf{0.833} $\pm$ 0.009 & 16.5 \\
         ${\dagger}$Latent-ODE-RNN \cite{rubanova2019latent} & 0.781 $\pm$ 0.018 & 6.7 \\
         ${\dagger}$Latent-ODE-ODE \cite{rubanova2019latent} & 0.829 $\pm$ 0.004 & 22.0 \\
         LTC \cite{hasani2021liquid} & 0.6477 $\pm$ 0.010 & 0.5 \\
         \hline
         Cf-S (ours) & 0.643 $\pm$ 0.018 & \textbf{0.1} \\
         CfC-noGate (ours) & \textbf{0.840} $\pm$ 0.003 & \textbf{0.1} \\
         CfC (ours) & \textbf{0.839} $\pm$ 0.002 & \textbf{0.1}  \\
         CfC-mmRNN (ours) & 0.834 +- 0.006 & \textbf{0.2} \\
         \bottomrule
    \end{tabular}
    \caption*{\footnotesize \textbf{Note:} The performance of the models marked by $\dagger$ are reported from  \cite{rubanova2019latent} and the ones with  $\ast$ from \cite{shukla2021multi}.}
    \label{tab:physionet}
\end{table}

\subsection*{Sentiment Analysis - IMDB}
The IMDB sentiment analysis dataset \cite{maas2011learning} consists of 25,000 training and 25,000 test sentences. Each sentence corresponds to either positive or negative sentiment. We tokenize the sentences in a word-by-word fashion with a vocabulary consisting of 20,000 most frequently occurring words in the dataset. We map each token to a vector using a trainable word embedding. The word embedding is initialized randomly. No pretraining of the network or the word embedding is performed. Table \ref{tab:imdb} represents how CfCs equipped with mixed memory instances outperform advanced RNN benchmarks. The hyperparameter details of this experiment is provided in Table \ref{tab:hyperparamsimdb}.

\begin{table}[t]
    \centering
    \caption{\textbf{Results on the IMDB datasets.} The experiment is performed without any pretraining or pretrained word-embeddings. Thus, we excluded advanced attention-based models \cite{shukla2021multi,ad-trans2021} such as Transformers \cite{vaswani2017attention} and RNN structures that use pretraining. Numbers present mean $\pm$ standard deviations, n=5}
    \begin{tabular}{lc}
    \toprule
         Model & Test accuracy (\%)  \\
    \midrule
          ${\dagger}$HiPPO-LagT \cite{gu2020hippo}  & \textbf{88.0} $\pm$ 0.2\\
           ${\dagger}$HiPPO-LegS \cite{gu2020hippo} & 88.0 $\pm$ 0.2\\
           ${\dagger}$LMU \cite{voelker2019legendre} & 87.7 $\pm$ 0.1\\
           ${\dagger}$LSTM \cite{hochreiter1997long}& 87.3 $\pm$ 0.4\\
           ${\dagger}$GRU \cite{chung2014empirical} & 86.2 $\pm$ n/a \\
          $\ast$ReLU GRU \cite{dey2017gate} &  84.8 $\pm$ n/a \\
          $\ast$Skip LSTM \cite{campos2017skip} &  86.6 $\pm$ n/a \\
           ${\dagger}$expRNN \cite{lezcano2019cheap} & 84.3 $\pm$ 0.3 \\
           ${\dagger}$Vanilla RNN \cite{campos2017skip} & 67.4 $\pm$ 7.7\\
          $\ast$coRNN \cite{rusch2021coupled} & 86.7 $\pm$ 0.3 \\
          LTC \cite{hasani2021liquid} & 61.8 $\pm$ 6.1 \\
    \midrule
        Cf-S (ours) & 81.7 $\pm$ 0.5\\
        CfC-noGate (ours) &  87.5 $\pm$ 0.1\\
        CfC (ours) & 85.9 $\pm$ 0.9\\
        CfC-mmRNN (ours) & \textbf{88.3} $\pm$ 0.1\\
    \bottomrule
    \end{tabular}
    \caption*{\textbf{Note:} The performance of the models marked by $\dagger$ are reported from \cite{gu2020hippo}, and $\ast$ are reported from \cite{rusch2021coupled}. The n/a standard deviation denotes that the original report of these experiments did not provide the statistics of their analysis.}
    \label{tab:imdb}
\end{table}

\subsection*{Physical Dynamics Modeling}
The Walker2D dataset consists of kinematic simulations of the MuJoCo physics engine \cite{todorov2012mujoco} on the \texttt{Walker2d-v2} OpenAI gym \cite{brockman2016openai} environment using four different stochastic policies.
The objective is to predict the physics state of the next time step. The training and testing sequences are provided at irregularly-sampled intervals. We report the squared error on the test set as a metric. As shown in Table \ref{tab:real_walker}, CfCs outperform the other baselines by a large margin rooting for their strong capability to model irregularly sampled physical dynamics with missing phases. It is worth mentioning that on this task, CfCs even outperform Transformers by a considerable 18\% margin. The hyperparameter details of this experiment is provided in Table \ref{tab:hyperparamswalker2d}. 

\begin{table}[t]
    \centering
     \caption{\textbf{Per time-step regression}. Modeling the physical dynamics of a Walker agent in simulation. Numbers present mean $\pm$ standard deviations. $n=5$}
     \begin{tabular}{llc}
     \toprule
     Model & Square-error & Time per epoch (min) \\
\hline
${\dagger}$ODE-RNN \cite{rubanova2019latent} & 1.904 $\pm$ 0.061 & 0.79 \\
${\dagger}$CT-RNN \cite{funahashi1993approximation} & 1.198 $\pm$ 0.004 & 0.91\\
${\dagger}$Augmented LSTM \cite{hochreiter1997long} & 1.065 $\pm$ 0.006 & 0.10 \\
${\dagger}$CT-GRU \cite{mozer2017discrete} & 1.172 $\pm$ 0.011  & 0.18 \\
${\dagger}$RNN-Decay \cite{rubanova2019latent} & 1.406 $\pm$ 0.005 & 0.16\\
${\dagger}$Bi-directional RNN \cite{schuster1997bidirectional} & 1.071 $\pm$ 0.009 & 0.39 \\
${\dagger}$GRU-D \cite{che2018recurrent}& 1.090 $\pm$ 0.034 & 0.11 \\
${\dagger}$PhasedLSTM \cite{neil2016phased} & 1.063 $\pm$ 0.010 & 0.25 \\
${\dagger}$GRU-ODE \cite{rubanova2019latent} & 1.051 $\pm$ 0.018 & 0.56 \\
${\dagger}$CT-LSTM \cite{mei2017neural} & 1.014 $\pm$ 0.014 & 0.31 \\
${\dagger}$ODE-LSTM \cite{lechner2020learning} & 0.883 $\pm$ 0.014 & 0.29 \\
coRNN \cite{rusch2021coupled} & 3.241 $\pm$ 0.215 & 0.18 \\
Lipschitz RNN \cite{erichson2021lipschitz} & 1.781 $\pm$ 0.013 & 0.17 \\
LTC \cite{hasani2021liquid} & \textbf{0.662} $\pm$ 0.013 & 0.78 \\
Transformer \cite{vaswani2017attention} & 0.761 $\pm$ 0.032 & 0.8 \\
     \cmidrule{1-3}
     Cf-S (ours) & 0.948  $\pm$ 0.009 & 0.12 \\
     CfC-noGate (ours) & \textbf{0.650}  $\pm$ 0.008 & 0.21 \\
     CfC (ours) & \textbf{0.643} $\pm$ 0.006 & 0.08 \\
     CfC-mmRNN (ours) & \textbf{0.617} $\pm$ 0.006 & 0.34\\
     \bottomrule
     \end{tabular}
    \caption*{\footnotesize \textbf{Note:} The performance of the models marked by $\dagger$ are reported from \cite{lechner2020learning}.}
 \label{tab:real_walker}
\end{table}

\section*{Scope, Discussions and Conclusions}
We introduced a closed-form continuous-time neural model build from an approximate close-form solution of liquid time-constant networks that possesses the strong modeling capabilities of ODE-based networks while being significantly faster, more accurate, and stable. These closed-form continuous-depth models achieve this by explicit time-dependent gating mechanisms and having a liquid time-constant modulated by neural networks.

\noindent \textbf{Continuous-Depth Models.} Machine learning, control theory and dynamical systems merge at models with continuous-time dynamics \cite{zhang2014comprehensive,weinan2017proposal,lu2017expressive,li2017maximum,lechner2019designing}. In a seminal work, Chen et. al. 2018 \cite{chen2018neural} revived the class of continuous-time neural networks \cite{cohen1983absolute,funahashi1993approximation}, with neural ODEs. These continuous-depth models give rise to vector field representations and a set of functions which were not possible to generate before with discrete neural networks. These capabilities enabled flexible density estimation \cite{grathwohl2018ffjord,dupont2019augmented,yang2019pointflow,NEURIPS2020_1aa3d9c6,hodgkinson2020stochastic}, as well as performant modeling of sequential and irregularly-sampled data \cite{rubanova2019latent,gholami2019anode,lechner2020learning,hasani2021liquid,erichson2021lipschitz}. In this paper, we showed how to relax the need for an ODE-solver to realize an expressive continuous-time neural network model for challenging time-series problems. 

\noindent \textbf{Improving Neural ODEs.} ODE-based neural networks are as good as their ODE-solvers. As the complexity or the dimensionality of the modeling task increases, ODE-based networks demand a more advanced solver that significantly impacts their efficiency \cite{poli2020hypersolvers}, stability  \cite{bai2019deep,haber2019imexnet,chang2019antisymmetricrnn,massaroli2020stable,lechner2020gershgorin} and performance \cite{hasani2021liquid}. A large body of research went into improving the computational overhead of these solvers, for example, by designing hypersolvers \cite{poli2020hypersolvers}, deploying augmentation methods \cite{dupont2019augmented,massaroli2020dissecting}, pruning \cite{liebenwein2021sparse} and by regularizing the continuous flows \cite{finlay2020train,massaroli2020stable,kidger2020hey}. To enhance the performance of an ODE-based model, especially in time series modeling tasks \cite{gleeson2018c302}, solutions provided for stabilizing their gradient propagation \cite{lechner2020learning,erichson2021lipschitz,li2020scalable}. In this work, we showed that CfCs improve the scalability, efficiency, and performance of continuous-depth neural models. 

\noindent \textbf{Now that we have a closed-form system, where does it make sense to use ODE-based networks?} For large-scale time-series prediction tasks, and where closed-loop performance matters \cite{vorbach2021causal} CfCs should be the method of choice.This is because, they capture the flexible, continuous-time nature of ODE-based networks while presenting large gains in performance and scalability. On the other hand, implicit ODE-based models can still be significantly beneficial in solving continuously defined physics problems and control tasks. Moreover, for generative modeling, continuous normalizing flows built by ODEs are the suitable choice of model as they ensure invertibility unlike CfCs \cite{chen2018neural}. This is because differential equations guarantee invertibility (i.e., under uniqueness conditions \cite{liebenwein2021sparse}, one can run them backwards in time). CfCs only approximate ODEs and therefore they no longer necessarily form a bijection \cite{rezende2015variational}. 

\noindent \textbf{What are the limitations of CfCs?} CfCs might express vanishing gradient problems. To avoid this, for tasks that require long-term dependencies, it is better to use them together with mixed memory networks \cite{lechner2020learning} (See CfC-mmRNN). Moreover, we speculate that inferring causality from ODE-based networks might be more straightforward than a closed-form solution \cite{vorbach2021causal}. It would also be beneficial to assess if verifying a continuous neural flow \cite{Grunbacher2021verification} is more tractable by an ODE representation of the system or their closed form. 

\noindent \textbf{In what application scenarios shall we use CfCs?} For problems such as language modeling where a significant amount of sequential data and substantial compute resources are available, Transformers \cite{vaswani2017attention} are the right choice. In contrast, we use CfCs when: 1) data has limitations and irregularities (e.g., medical data, financial time-series, robotics \cite{lechner2021adversarial} and closed loop control and robotics, and multi-agent autonomous systems in supervised and reinforcement learning schemes \cite{brunnbauer2021model}), 2) training and inference efficiency of a model is important (e.g., embedded applications \cite{hasani2016efficient,wang2019generative,delpreto2020plug}), and 3) when interpretability matters \cite{hasani2020interpretable}.


\begin{thebibliography}{10}
\expandafter\ifx\csname url\endcsname\relax
  \def\url#1{\texttt{#1}}\fi
\expandafter\ifx\csname urlprefix\endcsname\relax\def\urlprefix{URL }\fi
\providecommand{\bibinfo}[2]{#2}
\providecommand{\eprint}[2][]{\url{#2}}

\bibitem{hasani2021liquid}
\bibinfo{author}{Hasani, R.}, \bibinfo{author}{Lechner, M.},
  \bibinfo{author}{Amini, A.}, \bibinfo{author}{Rus, D.} \&
  \bibinfo{author}{Grosu, R.}
\newblock \bibinfo{title}{Liquid time-constant networks}.
\newblock \emph{\bibinfo{journal}{Proceedings of the AAAI Conference on
  Artificial Intelligence}} \textbf{\bibinfo{volume}{35}},
  \bibinfo{pages}{7657--7666} (\bibinfo{year}{2021}).

\bibitem{chen2018neural}
\bibinfo{author}{Chen, T.~Q.}, \bibinfo{author}{Rubanova, Y.},
  \bibinfo{author}{Bettencourt, J.} \& \bibinfo{author}{Duvenaud, D.~K.}
\newblock \bibinfo{title}{Neural ordinary differential equations}.
\newblock In \emph{\bibinfo{booktitle}{Advances in neural information
  processing systems}}, \bibinfo{pages}{6571--6583} (\bibinfo{year}{2018}).

\bibitem{grathwohl2018ffjord}
\bibinfo{author}{Grathwohl, W.}, \bibinfo{author}{Chen, R.~T.},
  \bibinfo{author}{Bettencourt, J.}, \bibinfo{author}{Sutskever, I.} \&
  \bibinfo{author}{Duvenaud, D.}
\newblock \bibinfo{title}{Ffjord: Free-form continuous dynamics for scalable
  reversible generative models}.
\newblock \emph{\bibinfo{journal}{arXiv preprint arXiv:1810.01367}}
  (\bibinfo{year}{2018}).

\bibitem{dupont2019augmented}
\bibinfo{author}{Dupont, E.}, \bibinfo{author}{Doucet, A.} \&
  \bibinfo{author}{Teh, Y.~W.}
\newblock \bibinfo{title}{Augmented neural odes}.
\newblock In \emph{\bibinfo{booktitle}{Advances in Neural Information
  Processing Systems}}, \bibinfo{pages}{3134--3144} (\bibinfo{year}{2019}).

\bibitem{yang2019pointflow}
\bibinfo{author}{Yang, G.} \emph{et~al.}
\newblock \bibinfo{title}{Pointflow: 3d point cloud generation with continuous
  normalizing flows}.
\newblock In \emph{\bibinfo{booktitle}{Proceedings of the IEEE/CVF
  International Conference on Computer Vision}}, \bibinfo{pages}{4541--4550}
  (\bibinfo{year}{2019}).

\bibitem{liebenwein2021sparse}
\bibinfo{author}{Liebenwein, L.}, \bibinfo{author}{Hasani, R.},
  \bibinfo{author}{Amini, A.} \& \bibinfo{author}{Daniela, R.}
\newblock \bibinfo{title}{Sparse flows: Pruning continuous-depth models}.
\newblock \emph{\bibinfo{journal}{arXiv preprint arXiv:2106.12718}}
  (\bibinfo{year}{2021}).

\bibitem{rubanova2019latent}
\bibinfo{author}{Rubanova, Y.}, \bibinfo{author}{Chen, R.~T.} \&
  \bibinfo{author}{Duvenaud, D.}
\newblock \bibinfo{title}{Latent odes for irregularly-sampled time series}.
\newblock \emph{\bibinfo{journal}{arXiv preprint arXiv:1907.03907}}
  (\bibinfo{year}{2019}).

\bibitem{gholami2019anode}
\bibinfo{author}{Gholami, A.}, \bibinfo{author}{Keutzer, K.} \&
  \bibinfo{author}{Biros, G.}
\newblock \bibinfo{title}{Anode: Unconditionally accurate memory-efficient
  gradients for neural odes}.
\newblock \emph{\bibinfo{journal}{arXiv preprint arXiv:1902.10298}}
  (\bibinfo{year}{2019}).

\bibitem{lechner2020learning}
\bibinfo{author}{Lechner, M.} \& \bibinfo{author}{Hasani, R.}
\newblock \bibinfo{title}{Learning long-term dependencies in
  irregularly-sampled time series}.
\newblock \emph{\bibinfo{journal}{arXiv preprint arXiv:2006.04418}}
  (\bibinfo{year}{2020}).

\bibitem{prince1981high}
\bibinfo{author}{Prince, P.~J.} \& \bibinfo{author}{Dormand, J.~R.}
\newblock \bibinfo{title}{High order embedded runge-kutta formulae}.
\newblock \emph{\bibinfo{journal}{Journal of computational and applied
  mathematics}} \textbf{\bibinfo{volume}{7}}, \bibinfo{pages}{67--75}
  (\bibinfo{year}{1981}).

\bibitem{raissi2019physics}
\bibinfo{author}{Raissi, M.}, \bibinfo{author}{Perdikaris, P.} \&
  \bibinfo{author}{Karniadakis, G.~E.}
\newblock \bibinfo{title}{Physics-informed neural networks: A deep learning
  framework for solving forward and inverse problems involving nonlinear
  partial differential equations}.
\newblock \emph{\bibinfo{journal}{Journal of Computational Physics}}
  \textbf{\bibinfo{volume}{378}}, \bibinfo{pages}{686--707}
  (\bibinfo{year}{2019}).

\bibitem{massaroli2020dissecting}
\bibinfo{author}{Massaroli, S.}, \bibinfo{author}{Poli, M.},
  \bibinfo{author}{Park, J.}, \bibinfo{author}{Yamashita, A.} \&
  \bibinfo{author}{Asma, H.}
\newblock \bibinfo{title}{Dissecting neural odes}.
\newblock In \emph{\bibinfo{booktitle}{34th Conference on Neural Information
  Processing Systems, NeurIPS 2020}} (\bibinfo{organization}{The Neural
  Information Processing Systems}, \bibinfo{year}{2020}).

\bibitem{bai2019deep}
\bibinfo{author}{Bai, S.}, \bibinfo{author}{Kolter, J.~Z.} \&
  \bibinfo{author}{Koltun, V.}
\newblock \bibinfo{title}{Deep equilibrium models}.
\newblock \emph{\bibinfo{journal}{Advances in Neural Information Processing
  Systems}} \textbf{\bibinfo{volume}{32}}, \bibinfo{pages}{690--701}
  (\bibinfo{year}{2019}).

\bibitem{finlay2020train}
\bibinfo{author}{Finlay, C.}, \bibinfo{author}{Jacobsen, J.-H.},
  \bibinfo{author}{Nurbekyan, L.} \& \bibinfo{author}{Oberman, A.~M.}
\newblock \bibinfo{title}{How to train your neural ode}.
\newblock \emph{\bibinfo{journal}{arXiv preprint arXiv:2002.02798}}
  (\bibinfo{year}{2020}).

\bibitem{massaroli2020stable}
\bibinfo{author}{Massaroli, S.} \emph{et~al.}
\newblock \bibinfo{title}{Stable neural flows}.
\newblock \emph{\bibinfo{journal}{arXiv preprint arXiv:2003.08063}}
  (\bibinfo{year}{2020}).

\bibitem{kidger2020hey}
\bibinfo{author}{Kidger, P.}, \bibinfo{author}{Chen, R.~T.} \&
  \bibinfo{author}{Lyons, T.}
\newblock \bibinfo{title}{" hey, that's not an ode": Faster ode adjoints with
  12 lines of code}.
\newblock \emph{\bibinfo{journal}{arXiv preprint arXiv:2009.09457}}
  (\bibinfo{year}{2020}).

\bibitem{poli2020hypersolvers}
\bibinfo{author}{Poli, M.} \emph{et~al.}
\newblock \bibinfo{title}{Hypersolvers: Toward fast continuous-depth models}.
\newblock \emph{\bibinfo{journal}{Advances in Neural Information Processing
  Systems}} \textbf{\bibinfo{volume}{33}} (\bibinfo{year}{2020}).

\bibitem{friston2003dynamic}
\bibinfo{author}{Friston, K.~J.}, \bibinfo{author}{Harrison, L.} \&
  \bibinfo{author}{Penny, W.}
\newblock \bibinfo{title}{Dynamic causal modelling}.
\newblock \emph{\bibinfo{journal}{Neuroimage}} \textbf{\bibinfo{volume}{19}},
  \bibinfo{pages}{1273--1302} (\bibinfo{year}{2003}).

\bibitem{PerkoODEs}
\bibinfo{author}{Perko, L.}
\newblock \emph{\bibinfo{title}{Differential Equations and Dynamical Systems}}
  (\bibinfo{publisher}{Springer-Verlag}, \bibinfo{address}{Berlin, Heidelberg},
  \bibinfo{year}{1991}).

\bibitem{hochreiter1997long}
\bibinfo{author}{Hochreiter, S.} \& \bibinfo{author}{Schmidhuber, J.}
\newblock \bibinfo{title}{Long short-term memory}.
\newblock \emph{\bibinfo{journal}{Neural computation}}
  \textbf{\bibinfo{volume}{9}}, \bibinfo{pages}{1735--1780}
  (\bibinfo{year}{1997}).

\bibitem{Rudin76}
\bibinfo{author}{Rudin, W.}
\newblock \emph{\bibinfo{title}{Principles of mathematical analysis}}
  (\bibinfo{publisher}{McGraw-Hill New York}, \bibinfo{year}{1976}),
  \bibinfo{edition}{3d ed.} edn.

\bibitem{lechner2020neural}
\bibinfo{author}{Lechner, M.} \emph{et~al.}
\newblock \bibinfo{title}{Neural circuit policies enabling auditable autonomy}.
\newblock \emph{\bibinfo{journal}{Nature Machine Intelligence}}
  \textbf{\bibinfo{volume}{2}}, \bibinfo{pages}{642--652}
  (\bibinfo{year}{2020}).

\bibitem{hochreiter1991untersuchungen}
\bibinfo{author}{Hochreiter, S.}
\newblock \bibinfo{title}{Untersuchungen zu dynamischen neuronalen netzen}.
\newblock \emph{\bibinfo{journal}{Diploma, Technische Universit{\"a}t
  M{\"u}nchen}} \textbf{\bibinfo{volume}{91}} (\bibinfo{year}{1991}).

\bibitem{vorbach2021causal}
\bibinfo{author}{Vorbach, C.}, \bibinfo{author}{Hasani, R.},
  \bibinfo{author}{Amini, A.}, \bibinfo{author}{Lechner, M.} \&
  \bibinfo{author}{Rus, D.}
\newblock \bibinfo{title}{Causal navigation by continuous-time neural
  networks}.
\newblock \emph{\bibinfo{journal}{arXiv preprint arXiv:2106.08314}}
  (\bibinfo{year}{2021}).

\bibitem{hasani2019response}
\bibinfo{author}{Hasani, R.} \emph{et~al.}
\newblock \bibinfo{title}{Response characterization for auditing cell dynamics
  in long short-term memory networks}.
\newblock In \emph{\bibinfo{booktitle}{2019 International Joint Conference on
  Neural Networks (IJCNN)}}, \bibinfo{pages}{1--8}
  (\bibinfo{organization}{IEEE}, \bibinfo{year}{2019}).

\bibitem{amini2021vista}
\bibinfo{author}{Amini, A.} \emph{et~al.}
\newblock \bibinfo{title}{Vista 2.0: An open, data-driven simulator for
  multimodal sensing and policy learning for autonomous vehicles}.
\newblock \emph{\bibinfo{journal}{arXiv preprint arXiv:2111.12083}}
  (\bibinfo{year}{2021}).

\bibitem{amini2020learning}
\bibinfo{author}{Amini, A.} \emph{et~al.}
\newblock \bibinfo{title}{Learning robust control policies for end-to-end
  autonomous driving from data-driven simulation}.
\newblock \emph{\bibinfo{journal}{IEEE Robotics and Automation Letters}}
  \textbf{\bibinfo{volume}{5}}, \bibinfo{pages}{1143--1150}
  (\bibinfo{year}{2020}).

\bibitem{levine2013guided}
\bibinfo{author}{Levine, S.} \& \bibinfo{author}{Koltun, V.}
\newblock \bibinfo{title}{Guided policy search}.
\newblock In \emph{\bibinfo{booktitle}{International conference on machine
  learning}}, \bibinfo{pages}{1--9} (\bibinfo{organization}{PMLR},
  \bibinfo{year}{2013}).

\bibitem{bojarski2018visualbackprop}
\bibinfo{author}{Bojarski, M.} \emph{et~al.}
\newblock \bibinfo{title}{Visualbackprop: Efficient visualization of cnns for
  autonomous driving}.
\newblock In \emph{\bibinfo{booktitle}{IEEE International Conference on
  Robotics and Automation (ICRA)}}, \bibinfo{pages}{1--8}
  (\bibinfo{year}{2018}).

\bibitem{chung2014empirical}
\bibinfo{author}{Chung, J.}, \bibinfo{author}{Gulcehre, C.},
  \bibinfo{author}{Cho, K.} \& \bibinfo{author}{Bengio, Y.}
\newblock \bibinfo{title}{Empirical evaluation of gated recurrent neural
  networks on sequence modeling}.
\newblock \emph{\bibinfo{journal}{arXiv preprint arXiv:1412.3555}}
  (\bibinfo{year}{2014}).

\bibitem{shukla2018interpolation}
\bibinfo{author}{Shukla, S.~N.} \& \bibinfo{author}{Marlin, B.}
\newblock \bibinfo{title}{Interpolation-prediction networks for irregularly
  sampled time series}.
\newblock In \emph{\bibinfo{booktitle}{International Conference on Learning
  Representations}} (\bibinfo{year}{2018}).

\bibitem{horn2020set}
\bibinfo{author}{Horn, M.}, \bibinfo{author}{Moor, M.}, \bibinfo{author}{Bock,
  C.}, \bibinfo{author}{Rieck, B.} \& \bibinfo{author}{Borgwardt, K.}
\newblock \bibinfo{title}{Set functions for time series}.
\newblock In \emph{\bibinfo{booktitle}{International Conference on Machine
  Learning}}, \bibinfo{pages}{4353--4363} (\bibinfo{organization}{PMLR},
  \bibinfo{year}{2020}).

\bibitem{funahashi1993approximation}
\bibinfo{author}{Funahashi, K.-i.} \& \bibinfo{author}{Nakamura, Y.}
\newblock \bibinfo{title}{Approximation of dynamical systems by continuous time
  recurrent neural networks}.
\newblock \emph{\bibinfo{journal}{Neural networks}}
  \textbf{\bibinfo{volume}{6}}, \bibinfo{pages}{801--806}
  (\bibinfo{year}{1993}).

\bibitem{mozer2017discrete}
\bibinfo{author}{Mozer, M.~C.}, \bibinfo{author}{Kazakov, D.} \&
  \bibinfo{author}{Lindsey, R.~V.}
\newblock \bibinfo{title}{Discrete event, continuous time rnns}.
\newblock \emph{\bibinfo{journal}{arXiv preprint arXiv:1710.04110}}
  (\bibinfo{year}{2017}).

\bibitem{mei2017neural}
\bibinfo{author}{Mei, H.} \& \bibinfo{author}{Eisner, J.}
\newblock \bibinfo{title}{The neural hawkes process: a neurally self-modulating
  multivariate point process}.
\newblock In \emph{\bibinfo{booktitle}{Proceedings of the 31st International
  Conference on Neural Information Processing Systems}},
  \bibinfo{pages}{6757--6767} (\bibinfo{year}{2017}).

\bibitem{che2018recurrent}
\bibinfo{author}{Che, Z.}, \bibinfo{author}{Purushotham, S.},
  \bibinfo{author}{Cho, K.}, \bibinfo{author}{Sontag, D.} \&
  \bibinfo{author}{Liu, Y.}
\newblock \bibinfo{title}{Recurrent neural networks for multivariate time
  series with missing values}.
\newblock \emph{\bibinfo{journal}{Scientific reports}}
  \textbf{\bibinfo{volume}{8}}, \bibinfo{pages}{1--12} (\bibinfo{year}{2018}).

\bibitem{neil2016phased}
\bibinfo{author}{Neil, D.}, \bibinfo{author}{Pfeiffer, M.} \&
  \bibinfo{author}{Liu, S.-C.}
\newblock \bibinfo{title}{Phased lstm: accelerating recurrent network training
  for long or event-based sequences}.
\newblock In \emph{\bibinfo{booktitle}{Proceedings of the 30th International
  Conference on Neural Information Processing Systems}},
  \bibinfo{pages}{3889--3897} (\bibinfo{year}{2016}).

\bibitem{schuster1997bidirectional}
\bibinfo{author}{Schuster, M.} \& \bibinfo{author}{Paliwal, K.~K.}
\newblock \bibinfo{title}{Bidirectional recurrent neural networks}.
\newblock \emph{\bibinfo{journal}{IEEE transactions on Signal Processing}}
  \textbf{\bibinfo{volume}{45}}, \bibinfo{pages}{2673--2681}
  (\bibinfo{year}{1997}).

\bibitem{voelker2019legendre}
\bibinfo{author}{Voelker, A.~R.}, \bibinfo{author}{Kaji{\'c}, I.} \&
  \bibinfo{author}{Eliasmith, C.}
\newblock \bibinfo{title}{Legendre memory units: Continuous-time representation
  in recurrent neural networks}.
\newblock \emph{\bibinfo{journal}{NeurIPS Reproducability Challenge}}
  (\bibinfo{year}{2019}).

\bibitem{gu2020hippo}
\bibinfo{author}{Gu, A.}, \bibinfo{author}{Dao, T.}, \bibinfo{author}{Ermon,
  S.}, \bibinfo{author}{Rudra, A.} \& \bibinfo{author}{R{\'e}, C.}
\newblock \bibinfo{title}{Hippo: Recurrent memory with optimal polynomial
  projections}.
\newblock \emph{\bibinfo{journal}{arXiv preprint arXiv:2008.07669}}
  (\bibinfo{year}{2020}).

\bibitem{lezcano2019cheap}
\bibinfo{author}{Lezcano-Casado, M.} \& \bibinfo{author}{Mart{\i}nez-Rubio, D.}
\newblock \bibinfo{title}{Cheap orthogonal constraints in neural networks: A
  simple parametrization of the orthogonal and unitary group}.
\newblock In \emph{\bibinfo{booktitle}{International Conference on Machine
  Learning}}, \bibinfo{pages}{3794--3803} (\bibinfo{organization}{PMLR},
  \bibinfo{year}{2019}).

\bibitem{rusch2021coupled}
\bibinfo{author}{Rusch, T.~K.} \& \bibinfo{author}{Mishra, S.}
\newblock \bibinfo{title}{Coupled oscillatory recurrent neural network
  (co{\{}rnn{\}}): An accurate and (gradient) stable architecture for learning
  long time dependencies}.
\newblock In \emph{\bibinfo{booktitle}{International Conference on Learning
  Representations}} (\bibinfo{year}{2021}).
\newblock \urlprefix\url{https://openreview.net/forum?id=F3s69XzWOia}.

\bibitem{erichson2021lipschitz}
\bibinfo{author}{Erichson, N.~B.}, \bibinfo{author}{Azencot, O.},
  \bibinfo{author}{Queiruga, A.}, \bibinfo{author}{Hodgkinson, L.} \&
  \bibinfo{author}{Mahoney, M.~W.}
\newblock \bibinfo{title}{Lipschitz recurrent neural networks}.
\newblock In \emph{\bibinfo{booktitle}{International Conference on Learning
  Representations}} (\bibinfo{year}{2021}).
\newblock \urlprefix\url{https://openreview.net/forum?id=-N7PBXqOUJZ}.

\bibitem{shukla2021multi}
\bibinfo{author}{Shukla, S.~N.} \& \bibinfo{author}{Marlin, B.~M.}
\newblock \bibinfo{title}{Multi-time attention networks for irregularly sampled
  time series}.
\newblock \emph{\bibinfo{journal}{arXiv preprint arXiv:2101.10318}}
  (\bibinfo{year}{2021}).

\bibitem{ad-trans2021}
\bibinfo{author}{Xiong, Y.} \emph{et~al.}
\newblock \bibinfo{title}{Nystr{\"{o}}mformer: {A} nystr{\"{o}}m-based
  algorithm for approximating self-attention}.
\newblock \emph{\bibinfo{journal}{CoRR}}
  \textbf{\bibinfo{volume}{abs/2102.03902}} (\bibinfo{year}{2021}).

\bibitem{vaswani2017attention}
\bibinfo{author}{Vaswani, A.} \emph{et~al.}
\newblock \bibinfo{title}{Attention is all you need}.
\newblock In \emph{\bibinfo{booktitle}{Advances in neural information
  processing systems}}, \bibinfo{pages}{5998--6008} (\bibinfo{year}{2017}).

\bibitem{maas2011learning}
\bibinfo{author}{Maas, A.} \emph{et~al.}
\newblock \bibinfo{title}{Learning word vectors for sentiment analysis}.
\newblock In \emph{\bibinfo{booktitle}{Proceedings of the 49th annual meeting
  of the association for computational linguistics: Human language
  technologies}}, \bibinfo{pages}{142--150} (\bibinfo{year}{2011}).

\bibitem{dey2017gate}
\bibinfo{author}{Dey, R.} \& \bibinfo{author}{Salem, F.~M.}
\newblock \bibinfo{title}{Gate-variants of gated recurrent unit (gru) neural
  networks}.
\newblock In \emph{\bibinfo{booktitle}{2017 IEEE 60th international midwest
  symposium on circuits and systems (MWSCAS)}}, \bibinfo{pages}{1597--1600}
  (\bibinfo{organization}{IEEE}, \bibinfo{year}{2017}).

\bibitem{campos2017skip}
\bibinfo{author}{Campos, V.}, \bibinfo{author}{Jou, B.},
  \bibinfo{author}{Gir{\'o}-i Nieto, X.}, \bibinfo{author}{Torres, J.} \&
  \bibinfo{author}{Chang, S.-F.}
\newblock \bibinfo{title}{Skip rnn: Learning to skip state updates in recurrent
  neural networks}.
\newblock \emph{\bibinfo{journal}{arXiv preprint arXiv:1708.06834}}
  (\bibinfo{year}{2017}).

\bibitem{todorov2012mujoco}
\bibinfo{author}{Todorov, E.}, \bibinfo{author}{Erez, T.} \&
  \bibinfo{author}{Tassa, Y.}
\newblock \bibinfo{title}{Mujoco: A physics engine for model-based control}.
\newblock In \emph{\bibinfo{booktitle}{2012 IEEE/RSJ International Conference
  on Intelligent Robots and Systems}}, \bibinfo{pages}{5026--5033}
  (\bibinfo{organization}{IEEE}, \bibinfo{year}{2012}).

\bibitem{brockman2016openai}
\bibinfo{author}{Brockman, G.} \emph{et~al.}
\newblock \bibinfo{title}{Openai gym}.
\newblock \emph{\bibinfo{journal}{arXiv preprint arXiv:1606.01540}}
  (\bibinfo{year}{2016}).

\bibitem{zhang2014comprehensive}
\bibinfo{author}{Zhang, H.}, \bibinfo{author}{Wang, Z.} \&
  \bibinfo{author}{Liu, D.}
\newblock \bibinfo{title}{A comprehensive review of stability analysis of
  continuous-time recurrent neural networks}.
\newblock \emph{\bibinfo{journal}{IEEE Transactions on Neural Networks and
  Learning Systems}} \textbf{\bibinfo{volume}{25}}, \bibinfo{pages}{1229--1262}
  (\bibinfo{year}{2014}).

\bibitem{weinan2017proposal}
\bibinfo{author}{Weinan, E.}
\newblock \bibinfo{title}{A proposal on machine learning via dynamical
  systems}.
\newblock \emph{\bibinfo{journal}{Communications in Mathematics and
  Statistics}} \textbf{\bibinfo{volume}{5}}, \bibinfo{pages}{1--11}
  (\bibinfo{year}{2017}).

\bibitem{lu2017expressive}
\bibinfo{author}{Lu, Z.}, \bibinfo{author}{Pu, H.}, \bibinfo{author}{Wang, F.},
  \bibinfo{author}{Hu, Z.} \& \bibinfo{author}{Wang, L.}
\newblock \bibinfo{title}{The expressive power of neural networks: A view from
  the width}.
\newblock \emph{\bibinfo{journal}{arXiv preprint arXiv:1709.02540}}
  (\bibinfo{year}{2017}).

\bibitem{li2017maximum}
\bibinfo{author}{Li, Q.}, \bibinfo{author}{Chen, L.}, \bibinfo{author}{Tai, C.}
  \emph{et~al.}
\newblock \bibinfo{title}{Maximum principle based algorithms for deep
  learning}.
\newblock \emph{\bibinfo{journal}{arXiv preprint arXiv:1710.09513}}
  (\bibinfo{year}{2017}).

\bibitem{lechner2019designing}
\bibinfo{author}{Lechner, M.}, \bibinfo{author}{Hasani, R.},
  \bibinfo{author}{Zimmer, M.}, \bibinfo{author}{Henzinger, T.~A.} \&
  \bibinfo{author}{Grosu, R.}
\newblock \bibinfo{title}{Designing worm-inspired neural networks for
  interpretable robotic control}.
\newblock In \emph{\bibinfo{booktitle}{International Conference on Robotics and
  Automation (ICRA)}}, \bibinfo{pages}{87--94} (\bibinfo{year}{2019}).

\bibitem{cohen1983absolute}
\bibinfo{author}{Cohen, M.~A.} \& \bibinfo{author}{Grossberg, S.}
\newblock \bibinfo{title}{Absolute stability of global pattern formation and
  parallel memory storage by competitive neural networks}.
\newblock \emph{\bibinfo{journal}{IEEE transactions on systems, man, and
  cybernetics}} \bibinfo{pages}{815--826} (\bibinfo{year}{1983}).

\bibitem{NEURIPS2020_1aa3d9c6}
\bibinfo{author}{Mathieu, E.} \& \bibinfo{author}{Nickel, M.}
\newblock \bibinfo{title}{Riemannian continuous normalizing flows}.
\newblock In \bibinfo{editor}{Larochelle, H.}, \bibinfo{editor}{Ranzato, M.},
  \bibinfo{editor}{Hadsell, R.}, \bibinfo{editor}{Balcan, M.~F.} \&
  \bibinfo{editor}{Lin, H.} (eds.) \emph{\bibinfo{booktitle}{Advances in Neural
  Information Processing Systems}}, vol.~\bibinfo{volume}{33},
  \bibinfo{pages}{2503--2515} (\bibinfo{publisher}{Curran Associates, Inc.},
  \bibinfo{year}{2020}).
\newblock
  \urlprefix\url{https://proceedings.neurips.cc/paper/2020/file/1aa3d9c6ce672447e1e5d0f1b5207e85-Paper.pdf}.

\bibitem{hodgkinson2020stochastic}
\bibinfo{author}{Hodgkinson, L.}, \bibinfo{author}{van~der Heide, C.},
  \bibinfo{author}{Roosta, F.} \& \bibinfo{author}{Mahoney, M.~W.}
\newblock \bibinfo{title}{Stochastic normalizing flows}.
\newblock \emph{\bibinfo{journal}{arXiv preprint arXiv:2002.09547}}
  (\bibinfo{year}{2020}).

\bibitem{haber2019imexnet}
\bibinfo{author}{Haber, E.}, \bibinfo{author}{Lensink, K.},
  \bibinfo{author}{Treister, E.} \& \bibinfo{author}{Ruthotto, L.}
\newblock \bibinfo{title}{Imexnet a forward stable deep neural network}.
\newblock In \emph{\bibinfo{booktitle}{International Conference on Machine
  Learning}}, \bibinfo{pages}{2525--2534} (\bibinfo{organization}{PMLR},
  \bibinfo{year}{2019}).

\bibitem{chang2019antisymmetricrnn}
\bibinfo{author}{Chang, B.}, \bibinfo{author}{Chen, M.},
  \bibinfo{author}{Haber, E.} \& \bibinfo{author}{Chi, E.~H.}
\newblock \bibinfo{title}{Antisymmetricrnn: A dynamical system view on
  recurrent neural networks}.
\newblock \emph{\bibinfo{journal}{arXiv preprint arXiv:1902.09689}}
  (\bibinfo{year}{2019}).

\bibitem{lechner2020gershgorin}
\bibinfo{author}{Lechner, M.}, \bibinfo{author}{Hasani, R.},
  \bibinfo{author}{Rus, D.} \& \bibinfo{author}{Grosu, R.}
\newblock \bibinfo{title}{Gershgorin loss stabilizes the recurrent neural
  network compartment of an end-to-end robot learning scheme}.
\newblock In \emph{\bibinfo{booktitle}{2020 IEEE International Conference on
  Robotics and Automation (ICRA)}}, \bibinfo{pages}{5446--5452}
  (\bibinfo{organization}{IEEE}, \bibinfo{year}{2020}).

\bibitem{gleeson2018c302}
\bibinfo{author}{Gleeson, P.}, \bibinfo{author}{Lung, D.},
  \bibinfo{author}{Grosu, R.}, \bibinfo{author}{Hasani, R.} \&
  \bibinfo{author}{Larson, S.~D.}
\newblock \bibinfo{title}{c302: a multiscale framework for modelling the
  nervous system of caenorhabditis elegans}.
\newblock \emph{\bibinfo{journal}{Philosophical Transactions of the Royal
  Society B: Biological Sciences}} \textbf{\bibinfo{volume}{373}},
  \bibinfo{pages}{20170379} (\bibinfo{year}{2018}).

\bibitem{li2020scalable}
\bibinfo{author}{Li, X.}, \bibinfo{author}{Wong, T.-K.~L.},
  \bibinfo{author}{Chen, R.~T.} \& \bibinfo{author}{Duvenaud, D.}
\newblock \bibinfo{title}{Scalable gradients for stochastic differential
  equations}.
\newblock In \emph{\bibinfo{booktitle}{International Conference on Artificial
  Intelligence and Statistics}}, \bibinfo{pages}{3870--3882}
  (\bibinfo{organization}{PMLR}, \bibinfo{year}{2020}).

\bibitem{rezende2015variational}
\bibinfo{author}{Rezende, D.} \& \bibinfo{author}{Mohamed, S.}
\newblock \bibinfo{title}{Variational inference with normalizing flows}.
\newblock In \emph{\bibinfo{booktitle}{International conference on machine
  learning}}, \bibinfo{pages}{1530--1538} (\bibinfo{organization}{PMLR},
  \bibinfo{year}{2015}).

\bibitem{Grunbacher2021verification}
\bibinfo{author}{Grunbacher, S.} \emph{et~al.}
\newblock \bibinfo{title}{On the verification of neural odes with stochastic
  guarantees}.
\newblock \emph{\bibinfo{journal}{Proceedings of the AAAI Conference on
  Artificial Intelligence}} \textbf{\bibinfo{volume}{35}},
  \bibinfo{pages}{11525--11535} (\bibinfo{year}{2021}).

\bibitem{lechner2021adversarial}
\bibinfo{author}{Lechner, M.}, \bibinfo{author}{Hasani, R.},
  \bibinfo{author}{Grosu, R.}, \bibinfo{author}{Rus, D.} \&
  \bibinfo{author}{Henzinger, T.~A.}
\newblock \bibinfo{title}{Adversarial training is not ready for robot
  learning}.
\newblock \emph{\bibinfo{journal}{arXiv preprint arXiv:2103.08187}}
  (\bibinfo{year}{2021}).

\bibitem{brunnbauer2021model}
\bibinfo{author}{Brunnbauer, A.} \emph{et~al.}
\newblock \bibinfo{title}{Model-based versus model-free deep reinforcement
  learning for autonomous racing cars}.
\newblock \emph{\bibinfo{journal}{arXiv preprint arXiv:2103.04909}}
  (\bibinfo{year}{2021}).

\bibitem{hasani2016efficient}
\bibinfo{author}{Hasani, R.~M.}, \bibinfo{author}{Haerle, D.} \&
  \bibinfo{author}{Grosu, R.}
\newblock \bibinfo{title}{Efficient modeling of complex analog integrated
  circuits using neural networks}.
\newblock In \emph{\bibinfo{booktitle}{2016 12th Conference on Ph. D. Research
  in Microelectronics and Electronics (PRIME)}}, \bibinfo{pages}{1--4}
  (\bibinfo{organization}{IEEE}, \bibinfo{year}{2016}).

\bibitem{wang2019generative}
\bibinfo{author}{Wang, G.}, \bibinfo{author}{Ledwoch, A.},
  \bibinfo{author}{Hasani, R.~M.}, \bibinfo{author}{Grosu, R.} \&
  \bibinfo{author}{Brintrup, A.}
\newblock \bibinfo{title}{A generative neural network model for the quality
  prediction of work in progress products}.
\newblock \emph{\bibinfo{journal}{Applied Soft Computing}}
  \textbf{\bibinfo{volume}{85}}, \bibinfo{pages}{105683}
  (\bibinfo{year}{2019}).

\bibitem{delpreto2020plug}
\bibinfo{author}{DelPreto, J.} \emph{et~al.}
\newblock \bibinfo{title}{Plug-and-play supervisory control using muscle and
  brain signals for real-time gesture and error detection}.
\newblock \emph{\bibinfo{journal}{Autonomous Robots}}
  \textbf{\bibinfo{volume}{44}}, \bibinfo{pages}{1303--1322}
  (\bibinfo{year}{2020}).

\bibitem{hasani2020interpretable}
\bibinfo{author}{Hasani, R.}
\newblock \emph{\bibinfo{title}{Interpretable Recurrent Neural Networks in
  Continuous-time Control Environments}}.
\newblock \bibinfo{type}{{PhD} dissertation}, \bibinfo{school}{Technische
  Universit\"at Wien} (\bibinfo{year}{2020}).

\end{thebibliography}

\section*{Acknowledgments}
Authors would like to thank Tsun-Hsuan Wang, Patrick Kao, Makram Chahine, Wei Xiao, Xiao Li, Lianhao Yin, and Yutong Ben for useful suggestions and testing out CfC models for confirmation of results across other domains. \noindent \textbf{Funding:} R.H. and D.R. are partially supported by Boeing and MIT. M.L. is supported in part by the Austrian Science Fund (FWF) under grant Z211-N23 (Wittgenstein Award). A.A. is supported by the National Science Foundation (NSF) Graduate Research Fellowship Program. M.T. is supported by the Poul Due Jensen Foundation, grant 883901. This research was partially sponsored by the United States Air Force Research Laboratory and the United States Air Force Artificial Intelligence Accelerator and was accomplished under Cooperative Agreement Number FA8750-19-2-1000. The views and conclusions contained in this document are those of the authors and should not be interpreted as representing the official policies, either expressed or implied, of the United States Air Force or the U.S. Government. The U.S. Government is authorized to reproduce and distribute reprints for Government purposes notwithstanding any copyright notation herein. This work was further supported by The Boeing Company and the Office of Naval Research (ONR) Grant N00014-18-1-2830. \textbf{Data and materials availability:} All data, code, and materials used in the analysis are openly available at \url{https://github.com/raminmh/CfC} under Apache 2.0 License, for purposes of reproducing and extending the analysis.

\section*{List of Supplementary materials}
\noindent Materials and Methods. \\
\noindent Tables S1 to S4.

\clearpage

\beginsupplement

\section*{Supplementary Materials}
Here, we provide all supplementary materials used in our analysis. 

\subsection*{Materials and Methods}
In this section, we provide the full proof for Lemma \ref{lemma:approx}.

\noindent \textbf{Proof of Lemma \ref{lemma:approx}}
\begin{proof}
We start by noting that
\begin{align*}
x(t) - \tilde{x}(t) & = c \big[ e^{- w_\tau t - \int_0^t f(I(s)) ds } - e^{- w_\tau t - f(I(t)) t }  f(-I(t)) \big] \\
& = c e^{- w_\tau t} \big[ e^{-\int_0^t f(I(s)) ds } - e^{- f(I(t)) t }  f(-I(t)) \big]
\end{align*}
Since $0 \leq f \leq 1$, we conclude $e^{-\int_0^t f(I(s)) ds } \in [0;1]$ and $e^{- f(I(t)) t } f(-I(t)) \in [0;1]$. This shows that $|x(t) - \tilde{x}(t)| \leq |c| e^{- w_\tau t}$. To see the sharpness results, pick some arbitrary small $\varepsilon > 0$ and a sufficiently large $C > 0$ such that $f(-C) \leq \varepsilon$ and $1 - \varepsilon \leq f(C)$. With this, for any $0 < \delta < t$, we consider the piecewise constant input signal $I$ such that $I(s) = -C$ for $s \in [0; t - \delta]$ and $I(s) = C$ for $s \in (t - \delta ; t]$. Then, it can be noted that
\begin{align*}
e^{-\int_0^t f(I(s)) ds } - e^{-f(I(t)) t} f(-I(t)) & \geq \\ 
e^{- \varepsilon t - \delta \cdot 1 } - e^{- (1 - \varepsilon) \cdot t} \varepsilon \to 1 , \quad \text{when} \ \ \varepsilon, \delta \to 0
\end{align*}

Statement 1) follows by noting that there exists a family of continuous signals $I_n : [0;t] \to \RE$ such that $|I_n(\cdot)| \leq C$ for all $n \geq 1$ and $I_n \to I$ pointwise as $n \to \infty$. This is because
\begin{align*}
\lim_{n \to \infty} \Big| \int_0^t f(I(s)) ds -  \int_0^t f(I_n(s)) ds \Big| \leq \\ \lim_{n \to \infty} \int_0^t |f(I(s)) - f(I_n(s))| ds \leq \lim_{n \to \infty} L \int_0^t |I(s) - I_n(s)| ds \\
= 0
\end{align*}
where $L$ is the Lipschitz constant of $f$ and the last identity is due to dominated convergence theorem~ \cite{Rudin76}. To see 2), we first note that the negation of the signal $-I$ provides us with
\begin{align*}
e^{-\int_0^t f(-I(s)) ds } - e^{-f(-I(t)) t} f(I(t)) \leq \\ e^{- (1 - \varepsilon) (t - \delta) - \delta \cdot 0 } - e^{- \varepsilon \cdot t} (1 - \varepsilon) \to e^{-t} - 1 ,
\end{align*}
if $\varepsilon, \delta \to 0$. The fact that the left-hand side of the last inequality must be at least $e^{-t} - 1$ follows by observing that $e^{-t} \leq e^{-\int_0^t f(I'(s)) ds } $ and $e^{-f(I''(t)) t} f(-I''(t)) \leq 1$ for any $I',I'' : [0;t] \to \RE$.
\end{proof}

\clearpage

\begin{table}[t]
    \centering
    \caption{\textbf{Bit-Stream XOR experiments}. Hyperparameters}
\begin{tabular}{lllll}
\toprule
Parameter & \multicolumn{4}{c}{Value}  \\
\midrule
{} & Cf-S & CfC & CfC-noGate & CfC-mmRNN \\
\midrule
clipnorm & 5 & 1 & 10 & 10 \\ 
optimizer & Adam & RMSProp & RMSprop & RMSprop \\
batch\_size & 256 & 128 & 128 & 128\\
Hidden size & 64 & 192 & 128 & 64 \\
epochs      & 200 & 200 & 200 & 200 \\ 
   base\_lr & 0.005 & 0.05 & 0.005 & 0.005\\
   decay\_lr & 0.9 & 0.7 & 0.95 & 0.95 \\ 
   backbone\_activation & SiLU & ReLU & SiLU & ReLU \\ 
   backbone\_dr & 0.0 & 0.0 & 0.3 & 0.0 \\
   forget\_bias & 1.2 & 1.2 & 4.7 & 0.6 \\
   backbone\_units & 64 & 128 & 192 & 128 \\
   backbone\_layers & 1 & 1 & 1 & 1 \\
   weight\_decay & 3e-05 & 3e-06 & 5e-06 & 2e-06 \\ 
\bottomrule
\end{tabular}
    \label{tab:hyperparamsxor}
\end{table}

\begin{table}[t]
    \centering
    \caption{\textbf{Physionet experiments}. Hyperparameters}
\begin{tabular}{lllll}
\toprule
Parameter & \multicolumn{4}{c}{Value}  \\
\midrule
{} & Cf-S & CfC & CfC-noGate & CfC-mmRNN \\
\midrule
epochs & 116 & 57 & 58 & 65 \\ 
class\_weight & 18.25 & 11.69 & 7.73 & 5.91 \\ 
clipnorm & 0 & 0 & 0 & 0 \\ 
Hidden size & 64 & 256 & 64 & 64 \\
   base\_lr & 0.003 & 0.002 & 0.003 & 0.001\\
   decay\_lr & 0.72 & 0.9 & 0.73 & 0.9 \\ 
   backbone\_activation & Tanh & SiLU & ReLU & LeCun Tanh \\ 
   backbone\_units & 64 & 64 & 192 & 64 \\
   backbone\_dr & 0.1 & 0.2 & 0.0 & 0.3 \\
   backbone\_layers & 3 & 2 & 2 & 2 \\
   weight\_decay & 5e-05 & 4e-06 & 5e-05 & 4e-06 \\ 
   optimizer & AdamW & AdamW & AdamW & AdamW \\
   init & 0.53 & 0.50 & 0.55 & 0.6 \\
   batch\_size & 128 & 128 & 128 & 128\\
\bottomrule
\end{tabular}
    \label{tab:hyperparamsphysionet}
\end{table}

\begin{table}[t]
    \centering
    \caption{\textbf{IMDB experiments}. Hyperparameters}
\begin{tabular}{lllll}
\toprule
Parameter & \multicolumn{4}{c}{Value}  \\
\midrule
{} & Cf-S & CfC & CfC-noGate & CfC-mmRNN \\
\midrule
clipnorm & 1 & 10 & 5 & 10 \\ 
optimizer & Adam & RMSProp & RMSprop & RMSprop \\
batch\_size & 128 & 128 & 128 & 128\\
Hidden size & 320 & 192 & 224 & 64 \\
   embed\_dim & 64 & 192 & 192 & 32 \\
   embed\_dr & 0.0 & 0.0 & 0.2 & 0.3 \\
   epochs & 27 & 47 & 37 & 20 \\ 
   base\_lr & 0.0005 & 0.0005 & 0.0005 & 0.0005\\
   decay\_lr & 0.8 & 0.7 & 0.8 & 0.8 \\ 
   backbone\_activation & Relu & SiLU & SiLU & LeCun Tanh \\ 
   backbone\_dr & 0.0 & 0.0 & 0.1 & 0.0 \\
   backbone\_units & 64 & 64 & 128 & 64 \\
   backbone\_layers & 1 & 2 & 1 & 1 \\
   weight\_decay & 0.00048 & 3.6e-05 & 2.7e-05 & 0.00029 \\ 
\bottomrule
\end{tabular}
    \label{tab:hyperparamsimdb}
\end{table}

\begin{table}[t]
    \centering
    \caption{\textbf{Walker2D experiments}. Hyperparameters}
\begin{tabular}{lllll}
\toprule
Parameter & \multicolumn{4}{c}{Value}  \\
\midrule
{} & Cf-S & CfC & CfC-noGate & CfC-mmRNN \\
\midrule
clipnorm & 10 & 1 & 1 & 10 \\ 
optimizer & Adam & Adam & Adam & Adam \\
batch\_size & 128 & 256 & 128 & 128\\
Hidden size & 256 & 64 & 256 & 128 \\
epochs      & 50 & 50 & 50 & 50 \\ 
   base\_lr & 0.006 & 0.02 & 0.008 & 0.005\\
   decay\_lr & 0.95 & 0.95 & 0.95 & 0.95 \\ 
   backbone\_activation & SiLU & SiLU & LeCun Tanh & LeCun Tanh\\ 
   backbone\_dr & 0.0 & 0.1 & 0.1 & 0.2 \\
   forget\_bias & 5.0 & 1.6 & 2.8 & 2.1 \\
   backbone\_units & 192 & 256 & 128 & 128 \\
   backbone\_layers & 1 & 1 & 1 & 2 \\
   weight\_decay & 1e-06 & 1e-06 & 3e-05 & 6e-06 \\ 
\bottomrule
\end{tabular}
    \label{tab:hyperparamswalker2d}
\end{table}

\clearpage

\end{document}